%% file: main.tex
\documentclass[manuscript,screen,review=false]{acmart}

\usepackage{multirow}
\usepackage[ruled, linesnumbered]{algorithm2e}
\usepackage{xcolor}
\usepackage{dsfont}
\usepackage{bbm}
 
\newcommand{\mycomment}[1]{\textcolor{gray}{\textit{#1}}}

\AtBeginDocument{%
  \providecommand\BibTeX{{%
    \normalfont B\kern-0.5em{\scshape i\kern-0.25em b}\kern-0.8em\TeX}}}

\setcopyright{acmlicensed}
\copyrightyear{2018}
\acmYear{2018}
\acmDOI{}

\acmConference[arXiv]{Make sure to enter the correct
  conference title from your rights confirmation emai}{preprint}{}

\begin{document}

\title{FIT-RAG: Black-Box RAG with Factual Information and Token Reduction}

\author{Yuren Mao}
\email{yuren.mao@zju.edu.cn}
\author{Xuemei Dong}
\email{webmaster@marysville-ohio.com}
\author{Wenyi Xu}
\email{webmaster@marysville-ohio.com}
\author{Yunjun Gao}
\email{gaoyj@zju.edu.cn}
\author{Bin Wei}
\email{binwei@zju.edu.cn}
\affiliation{%
  \institution{Zhejiang University}
  \streetaddress{38 Zheda Road}
  \city{Hangzhou}
  \state{Zhejiang}
  \country{China}
  \postcode{310007}
}

\author{Ying Zhang}
\affiliation{%
  \institution{Zhejiang Gongshang University}
  \streetaddress{18 Xuezheng Rd}
  \city{Hangzhou}
  \country{China}}
\email{larst@affiliation.org}

\begin{abstract}

Due to the extraordinarily large number of parameters, fine-tuning Large Language Models (LLMs) to update long-tail or out-of-date knowledge is impractical in lots of applications. To avoid fine-tuning, we can alternatively treat a LLM as a black-box (i.e., freeze the parameters of the LLM) and augment it with a Retrieval-Augmented Generation (RAG) system, namely black-box RAG. Recently, black-box RAG has achieved success in knowledge-intensive tasks and has gained much attention. Existing black-box RAG methods typically fine-tune the retriever to cater to LLMs' preferences and concatenate all the retrieved documents as the input, which suffers from two issues: (1) Ignorance of  Factual Information. The LLM preferred documents may not contain the factual information for the given question, which can mislead the retriever and hurt the effectiveness of black-box RAG; (2) Waste of Tokens. Simply concatenating all the retrieved documents  brings large amounts of unnecessary tokens for LLMs, which degenerates the efficiency of black-box RAG. To address these issues, this paper proposes a novel black-box RAG framework which utilizes the factual information in the retrieval and reduces the number of tokens for augmentation, dubbed FIT-RAG. FIT-RAG utilizes the factual information by constructing a bi-label document scorer which takes the factual information and LLMs' preferences as labels respectively. Besides, it reduces the tokens by introducing a self-knowledge recognizer and a sub-document-level token reducer, which enables FIT-RAG to avoid unnecessary augmentation and reduce augmentation tokens as much as possible. FIT-RAG achieves both superior effectiveness and efficiency, which is validated by extensive experiments  across three open-domain question-answering datasets: TriviaQA, NQ and PopQA. FIT-RAG can improve the answering accuracy  of Llama2-13B-Chat by 14.3\% on TriviaQA, 19.9\% on NQ and 27.5\% on PopQA, respectively. Furthermore, it can save approximately half of the tokens on average across the three datasets. 
\end{abstract}



\maketitle

\input{text/1.Introduction}

\input{text/2.Related_work}

\input{text/3.Preliminaries}
\input{text/4.Methodology}

\input{text/5.Experiments}
\input{text/6.Conclusions}

\bibliography{reference}
\bibliographystyle{ACM-Reference-Format}

\end{document}

%% file: text/1.Introduction.tex
\section{INTRODUCTION}

Large language models (LLMs), which typically have billions of parameters, have demonstrated remarkable performance on a wide range of natural language processing tasks~\cite{brown2020language, kaplan2020scaling, roberts2020much}. Pretrained on massive text corpora, recent LLMs, such as GPT-4~\cite{gpt4}, showcase a level of sophistication that approaches human-like proficiency especially in text generation. 
However, the knowledge stored in the parameters of LLMs is fixed, which is susceptible to becoming out-of-date and disable LLMs to address tasks requiring time-sensitive information. Moreover, LLMs struggle to learn long-tail knowledge which appears infrequently in their training data~\cite{kandpal2023large, mallen2023not}. Additionally, due to the extraordinarily large number of parameters, frequent fine-tuning of LLMs to update knowledge is expensive and infeasible in practice. Out-of-date and long-tail knowledge lead to LLMs struggling with hallucinations and factual errors, especially in knowledge-intensive tasks.

\begin{figure}[t]
  \centering
  \includegraphics[width=0.96\linewidth]{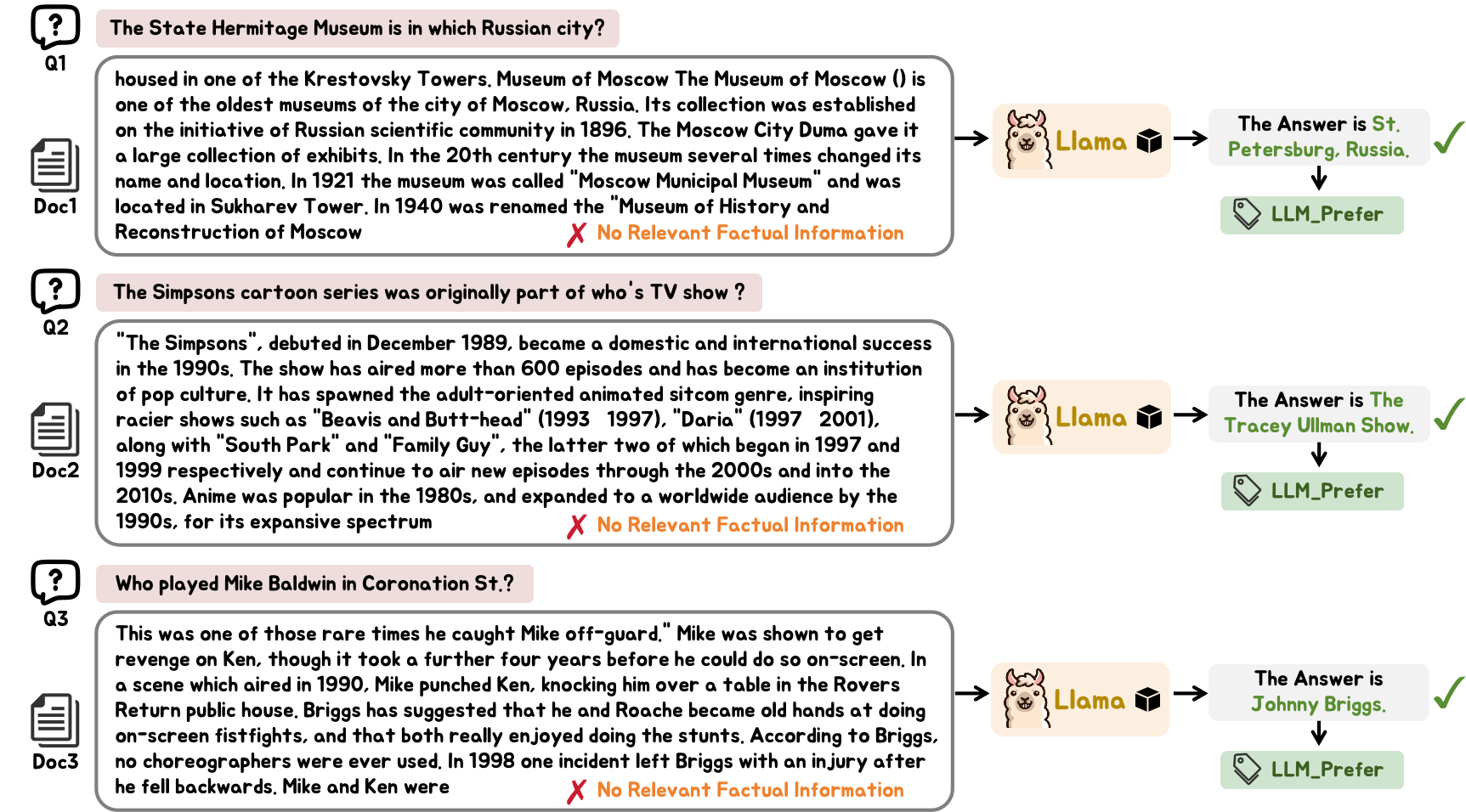}
  \caption{Examples illustrating LLM preferred retrieved documents that do not contain relevant factual information. These examples are obtained from the training set of TriviaQA and the answers are generated using Llama1-13B-Chat.}
  \label{fig:case}
\end{figure}

To address these issues, an emerging approach is Retrieval-Augmented Generation (RAG)~\cite{Khandelwal2020Generalization, retro, atlas, multimodalrag}. Instead of relying solely on LLM's inherent knowledge, RAG augments LLMs with external knowledge retrieved from large corpora or knowledge databases. Such RAG systems provide relevant context to ground the LLM's predictions and fill knowledge gaps. Prior RAG systems typically train both the retriever and the generation model to align with each other and adjust to downstream tasks~\cite{atlas, izacard2021distilling}. This joint training helps the generation model better utilize the retrieved information, and improves model synergy and generalization performance. However, this approach becomes impractical when the generation module is a large language model, which can have billions of parameters. On one hand, fine-tuning the full LLM is often infeasible due to the massive computational resources required; on the other hand, many existing LLMs are only accessible via APIs~\cite{ouyang2022training, gpt4} and cannot be fine-tuned.

To overcome the infeasibility of fine-tuning LLMs in RAG,  black-box RAG, which alternatively regards a LLM as a black-box (i.e., freeze the parameters of the LLM) and augment it without fine-tuning,  has achieved success in knowledge-intensive tasks and gained much attention. Existing black-box RAG methods~\cite{replug, aar, refeed, self-knowledge} typically fine-tune the retriever only based on LLMs' preferences (e.g., whether LLMs can give correct answer with the retrieved documents) and concatenate all the retrieved documents as the input, which suffers both effectiveness and efficiency issues. Only considering LLMs' preferences in retrieval causes \textbf{ignorance of factual information}, which can degenerate the effectiveness of RAG for it may mislead the retriever. As demonstrated in Figure~\ref{fig:case}, the LLM can answer correctly with the retrieved documents, but the documents themselves do not actually contain relevant factual information for the given question. For example, Q1 asks the location of the State Hermitage Museum; however, the retrieved document provides information about the Museum of Moscow. Although the LLM can give the correct answer, the retrieved document is actually unnecessary. If these unnecessary documents is used to reward the retriever, they can mislead the retriever. Besides, concatenating all the retrieved documents as the input  causes \textbf{waste of tokens}, which can introduce excessive unnecessary tokens and hurts the efficiency of RAG.

\begin{figure}[t]
  \centering
  \includegraphics[width=\linewidth]{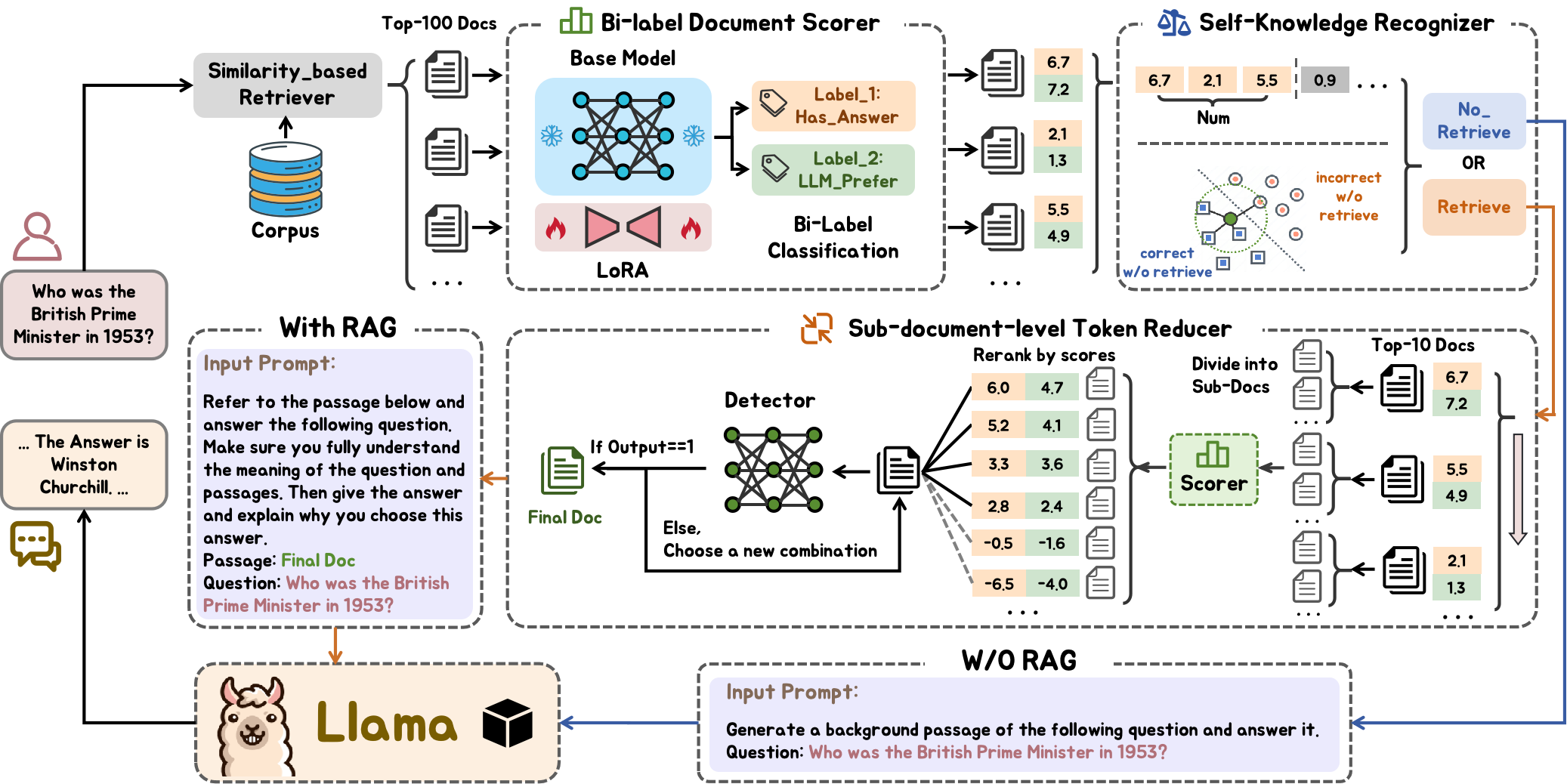}
  \caption{The overview of FIT-RAG}
  \label{fig:overview}
  \Description{}
\end{figure}

To simultaneously avoid the ignorance of factual information and the waste of tokens, this paper proposes a novel black-box RAG framework which utilizes both the factual information and LLM preference in the retrieval and performs token reduction for input tokens, dubbed FIT-RAG. Figure~\ref{fig:overview} gives the overall overview of FIT-RAG, which consists of five components: a similarity-based retriever, a bi-label document scorer, a bi-faceted self-knowledge recognizer, a sub-document-level token reducer and a prompt construction module. Among these components, the bi-label document scorer is proposed to effectively model alignment with LLM preferences as well as factual information, which avoids the ignorance of factual information; besides,  the bi-faceted self-knowledge recognizer and sub-document-level token reducer are proposed to reduce input tokens, which avoids the  waste of tokens.

The bi-label document scorer is learnt based on bi-label learning which includes a factual information label (\textit{Has\_Answer}) and a LLM preference label  (\textit{LLM\_Prefer}). The factual information label indicates whether the document containing the answer to the question, while the LLM preference label indicates whether the document help the LLM generate an accurate response. However, there is a serious data imbalance between the labels, which can degenerate the performance of  bi-label learning. To address the data imbalance, this paper proposes a data-imbalance-aware bi-label learning method, which allocates different weights for the different data, and the weights are automatically learnt with hypergradient-descent\cite{conf/iclr/BaydinCMSW18}. The proposed method can properly solve the data imbalance problem, and the bi-label document scorer can give a comprehensive evaluation for the retrieved documents.

The bi-faceted self-knowledge recognizer reduces input tokens by avoiding unnecessary augmentation, while the sub-document-level token reducer reduces input tokens by eliminating the unnecessary sub-documents.  The bi-faceted self-knowledge recognizer determines whether the LLM requires external knowledge by estimating whether the LLM has self-knowledge from two facet: whether the question is related to long-tail or out-of-date knowledge and whether the question’s nearest neighbors has self-knowledge. Besides, the sub-document-level token reducer eliminates the unnecessary sub-documents by selecting sub-documents combinations from the retrieved documents that have few sub-documents but is eligible to augment the LLM to give correct answers.

To verify the effectiveness of FIT-RAG, we adopt it to augment the Llama2-13B-Chat model on three open domain question answering datasets, TriviaQA, NQ and PopQA datasets, respectively. Compared with the original model Llama2-13B-Chat without retrieval augmentation, FIT-RAG improves the  answering  accuracy by 14.3\% on TriviaQA dataset, 19.9\% on NQ dataset and 27.5\% on PopQA dataset, respectively. Futhermore, it outperforms all other baseline RAG frameworks, which experimentally demonstrates the effectiveness of our proposed method.  Besides, FIT-RAG  consumes the least number of input tokens compared to all baseline black-box RAG methods. On average across the datasets, our proposed method can save approximately half of tokens, which greatly improves token efficiency and save computational resources.

%% file: text/2.Related_work.tex
\section{RELATED WORK}

\subsection{Large Language Models}
Recently, Large Language Models (LLMs) have grown rapidly in scale and capabilities. Early language models like BERT~\cite{bert} and T5~\cite{t5}, show strong performance on natural language understanding and generation tasks. These early successes spur further expansion of LLMs to even larger scales. Models such as InstructGPT~\cite{ouyang2022training}, LLama~\cite{llama}, OPT~\cite{opt} and BLOOM~\cite{bloom} processes parameters of tens or even hundreds of billions. This substantial increase in scale brings about a significant enhancement in the model's capacity. Recent models like GPT-4~\cite{gpt4}, which possesses an even larger scale, showcases a level of sophistication that approaches human-like proficiency.However, even the strongest GPT-4 model suffers from hallucinations and factual errors as the knowledge stored in the parameters is limited and easy to be out-of-date. To address these issues, a possible solution is Retrieval-Augmented Generation (RAG), which augments LLMs with external knowledge. Traditional RAG frameworks often target models with white-box settings, which may not be accessible in many scenarios since fine-tuning the full LLM requires massive computational resources and many LLMs can only be accessed through APIs~\cite{ouyang2022training,gpt4}. Therefore, we need to investigate RAG systems tailored for LLMs under black-box settings.

\subsection{Retrieval-Augmented Generation}

Retrieval-Augmented Generation (RAG) is a technique that augments natural language generation models with relevant content retrieved from knowledge sources, aiming at improving the quality and relevance of text generation. Previous works have demonstrated its strong performance in knowledge-intensive tasks such as question answering, fact checking, and content recommendation~\cite{kilt, promptcap, rag, factchecking}.

Retrievers interact with external corpus to acquire relevant information. For open-domain question answering, the Wikipedia corpus~\cite{wiki} is commonly used. As for retrieval methods, it can broadly be categorized into two types: sparse retrievers and dense retrievers. Sparse retrievers, such as TF-IDF~\cite{tfidf} and BM25~\cite{BM25}, predominantly rely on keyword matching for document retrieval. These methods determine the relevance between queries and documents by analyzing the occurrence and distribution of keywords within the documents. Dense retrievers employ dual-encoders to generate dense vector representations of text for more accurate semantic matching. Consequently, dense retrievers are considered more suitable for retrieval-augmented applications. Some techniques like vector quantization~\cite{Distill-VQ,LibVQ} and embedding optimization~\cite{emb-optim} also improves the efficiency of dense retrievers. Common dense retrievers include DPR~\cite{dpr}, ANCE~\cite{ance} and Contriever~\cite{contriever}. Specifically, DPR~\cite{dpr} is trained with supervised learning on question-answer pairs, and focuses on extracting relevant passages by analyzing the semantic content of both questions and answers. ANCE~\cite{ance} leverages approximate nearest neighbor search and contrastive learning to enhance the model's ability to discern between relevant and non-relevant documents in a dense vector space. Contriever~\cite{contriever} employs unsupervised contrastive learning to adapt to inherent data structure, especially beneficial when the annotated training data is scarce.
To enhance the quality of retrieved documents, some work conduct further reranking to these documents for personalization~\cite{teevan2005personalizing,bennett2012modeling,zhou2021pssl} and diversification~\cite{liu2020dvgan,su2021modeling}.

Recent work has explored different ways for language models to leverage retrieved or generated text as external knowledge. One approach is to integrate retrieval into language model pre-training or fine-tuning. For instance, REALM~\cite{realm} integrates external document retrieval into pre-training, enhancing performance in downstream tasks by retrieving relevant information. RAG~\cite{rag} adopts a generative approach that blends retrieval and generation. It is specifically fine-tuned for knowledge-intensive tasks like open-domain question answering, leveraging the synergy of retrieval-augmented capabilities with generative modeling. Atlas~\cite{atlas} extends upon the RAG framework by combining RAG's retrieval-generation method with encoder-decoder language model pre-training, with a focus on fine-tuning for question answering tasks. Another approach RETRO~\cite{retro} modifies the language model architecture for better text retrieval by employing kNN-LM to retrieve contextually relevant tokens and integrating their distribution with the predictions of the language model.

Instead of retrieval, some methods use text generation as the knowledge source. The concept is that knowledge within the language model can be \textit{retrieved} through direct text generation~\cite{lmknow}. For example, the Selfmem framework~\cite{selfmem} ingeniously employs a generator to repeatedly produce synthesized texts, forming an unlimited \textit{memory pool} for future use. This model uniquely uses its own generated outputs as \textit{self-memory} to aid subsequent generation, showcasing an innovative approach where text generation helps fabricate new memory references. Another notable method is RECITE~\cite{recite}, which is designed to enable LLMs to produce relevant information without resorting to external data sources. It first prompts the LLM to \textit{recite} relevant passages based on its internal knowledge. These passages are then used as a pseudo evidence document that the LLM conditions on to produce the final answer. Similarly, GENREAD~\cite{genread} also utilizes the generative capabilities of LLMs to avoid external retrieval, prompting the LLM to generate context-specific documents in response to the question. The LLM then reads these synthesized documents and generates the final response.

\subsection{Retrieval-Augmentation for Black-box Languages Models}

Large language models, such as InstructGPT~\cite{ouyang2022training} and GPT-4~\cite{gpt4}, are often non-open-source and exist as black-box APIs, allowing users only to send queries and receive responses without access or modification to their internal parameters. Traditional retrieval-augmented models typically focus on a white-box setup that is tunable, but this is infeasible for large-scale black-box language models. Addressing this challenge, recent research has developed retrieval augmentation  methods suitable for black-box settings. For instance, REPLUG~\cite{replug} operates by utilizing a fixed language model to evaluate and provide probability distributions for the documents retrieved. This supervises the retriever to select documents preferred by the LLM. Another method, AAR~\cite{aar}, creates positive and negative documents for a given question and uses these documents to fine-tune the retriever to align with the LLM's preference. REFEED~\cite{refeed} first creates answers, then uses a retrieval model to obtain relevant information from large document corpus based on the question and answers, and finally integrates the retrieved information into the in-context demonstration for output refinement. LRT~\cite{lrt} addresses the high computational cost issue in updating databases by introducing an adaptive similarity matching module and fine-tuning with fewer than one million parameters. Despite the applicability of the above-mentioned methods for retrieval augmentation in black-box settings, these methods typically ignore the importance of factual information and face issues of input token inefficiency, which hurts both the effectiveness and efficiency of the RAG system.

%% file: text/3.Preliminaries.tex
\section{PRELIMINARIES}

\subsection{Problem Formulation}

This paper focuses on Retrieval-Augmented Generation (RAG) system for black-box Large Language Models (LLMs), namely black-box RAG. In this section, we first give the definition of RAG and subsequently introduce the black-box RAG.

\textbf{Retrieval-Augmented Generation (RAG).} 
Given a natural language question $q$, an external knowledge corpus $\mathcal{W}$ and a generative language model $\mathcal{M}$, a RAG system aims to help $\mathcal{M}$ generate more accurate and informative responses to $q$ using a retrieval model $\mathcal{R}$, which effectively retrieves relevant documents $\mathcal{D} = ({d}_{1}, {d}_{2}, {d}_{3}, ...)$ from $\mathcal{W}$. The form of introducing external knowledge to the language model varies, including modifying attention weights during generation, incorporating it into input prompts, or using it in post-calibration of the model output. Moreover, existing RAG methods typically require joint fine-tuning of the retriever and the language model (e.g. Atlas~\cite{atlas}, REALM~\cite{realm}). However, joint fine-tuning is unaffordable in amount of  practical scenarios due to the extremely large parameter scale of LLMs. In these scenarios, we can alternatively treat an LLM as a black-box (i.e., freeze the parameters of the LLM) and augment it with a RAG system, namely black-box RAG. Next, we introduce the definition of the black-box RAG.

\textbf{Retrieval-Augmented Generation System for Black-box LLM (Black-box RAG).}
A RAG system for black-box LLM aims to enhance the generation capability of the black-box LLM $\mathcal{M}_{B}$ by retrieving external knowledge without updating the LLM parameters. While the parameters of the black-box LLM $\mathcal{M}_{B}$ are frozen, the parameters of the retrieval model $\mathcal{R}$ are learnable. Thus, the RAG system for black-box LLM only optimizes $\mathcal{R}$ to improve overall system performance, without modifying $\mathcal{M}_{B}$. Moreover, existing black-box RAG systems typically inject the retrieved documents $\mathcal{D}$ into $\mathcal{M}_{B}$ by constructing an input prompt that concatenates the question $q$ and documents $\mathcal{D}$, which leverages the powerful in-context learning capabilities of the LLM.

\subsection{Parameter-Efficient Fine-Tuning	}
Parameter-Efficient Fine-Tuning (PEFT) methods, which fine-tune only a small or additional set of model parameters and keep the majority of pre-trained parameters fixed, can largely reduce the computational cost of model adaptation. This makes PEFT more practical compared to full parameter fine-tuning, especially for large language models. Recently, various PEFT methods have been proposed, such as  Adapter~\cite{adapter}, Prompt Tuning~\cite{prompt_tuning}, Prefix-Tuning~\cite{prefix_tuning} and LoRA~\cite{LoRA}. These methods have shown competitive performance compared to full parameter fine-tuning on various downstream tasks. In our framework, we employ LoRA method to fine-tune our T5-based bi-label document scorer. Specifically, LoRA fine-tunes the model by introducing an additional low-rank matrix. By optimizing only the parameters constructing the low-rank matrix, LoRA adapts T5 to the downstream task while keeping the original T5 parameters frozen, which greatly reduces the computational cost and keeps competitive performance.
 
\subsection{Hallucination of LLMs}
The hallucination of LLMs refers to the phenomenon where LLMs generate content that seems reasonable but is actually incorrect, irrelevant to the input prompt, or even self-contradictory. Despite their impressive capabilities, LLMs still face challenges of hallucination. The hallucination phenomenon in LLMs is closely tied to their uncertainty and overconfidence. Without awareness of what they do not know, LLMs tend to exhibit excessive faith in their own predictions, oblivious to potential knowledge gaps. Addressing hallucination in LLMs is crucial for ensuring their reliability in real-world applications. 
In this work, we aim to alleviate the potential for hallucination in LLMs by augmenting them with external knowledge using our proposed FIT-RAG framework. By providing relevant factual information from knowledge sources, we ground the generation of LLMs in tangible evidence, enhancing their capacity for accurate and contextually relevant outputs.

\subsection{Prompt Learning}
Prompt learning is an emerging technique which stimulates the capabilities of LLMs through  designing proper prompts. Unlike traditional fine-tuning, where model parameters are directly updated, this approach emphasizes the crucial role of well-crafted prompts in guiding language models to produce more accurate and contextually relevant outputs. Moreover, by constructing effective prompts, LLMs can be customized for new tasks without costly training. This makes prompt learning much more practical and applicable in various natural language processing domains.
In this work, we find that different prompts significantly impact the ability of LLMs to leverage external information and influence the performance of the generated outputs. By carefully designing prompts tailored for our RAG system, the generation quality of the LLM is greatly improved.

%% file: text/4.Methodology.tex
\section{METHODOLOGY}

\begin{algorithm}
\caption{Inference of FIT-RAG}
\label{alg:infer}

\textbf{Required:} a given Question $q$, the Corpus $\mathcal{W}$, the Similarity-based Retriever $\mathcal{R}$, the Bi-label Document Scorer $\mathcal{S}$, the Self-Knowledge Recognizer $\mathcal{K}$, the Token Reducer $\mathcal{T}$, the Large Language Model $\mathcal{M}$. \\
$\mathcal{D}_{r} \gets$ use $\mathcal{R}$ to retrieve Top-100 relevant text documents from $\mathcal{W}$ given ${q}$ \mycomment{\tcp*{Similarity-Based Retriever}}

\For{$d_{r} \in \mathcal{D}_{r}$} {
    $( score_{1}$, $score_{2} ) \gets$ use $\mathcal{S}$ to generate bi-label scores of $d_r$ \mycomment{\tcp*{Bi-label Document Scorer}}
} 

$S_{ltod}(q) \gets$ measure the relevance to the long-tail knowledge or out-of-date knowledge of ${q}$\\
$S_{nn}(q) \gets$ measure the self-knowledge of the question’s nearest neighbors of ${q}$\\
$\mathcal{K}(q) \gets$ decide whether retrieval is necessary for ${q}$ using $\mathcal{K}$ according to $S_{ltod}$ and $S_{nn}$ 
\mycomment{\tcp*{Bi-faceted Self-Knowledge Recognizer}}
\If{$\mathcal{K}(q) == \textcolor{blue}{\textbf{No\_ Retrieve}} $}{
    $\mathcal{P}({q}) \gets$ construct the input prompt using only ${q}$ \mycomment{\tcp*{Prompt Construction}}
    $answer \gets$ use $\mathcal{M}$ to generate the answer given $\mathcal{P}({q})$\\
}
\ElseIf{$\mathcal{K}(q) == \textcolor{orange}{\textbf{Retrieve}} $}{

    $\mathcal{D}_f \gets$ Get the compressed sub-documents combination using $\mathcal{T}$( $\mathcal{D}_{r}$ )  \mycomment{\tcp*{Sub-document-level Token Reducer}}
    $\mathcal{P}({q}, \mathcal{D}_f) \gets$ construct the input prompt using ${q}$ and $\mathcal{D}_f$ \mycomment{\tcp*{Prompt Construction}}
    $answer \gets$ use $\mathcal{M}$ to generate the answer given $\mathcal{P}(q, \mathcal{D}_f)$\\
}
\end{algorithm}

\subsection{Overall Framework of FIT-RAG}
In this section, we present an overview of FIT-RAG, which simultaneously achieves effectiveness and efficiency based on our proposed  bi-label document scorer, bi-faceted self-knowledge recognizer and sub-document-level token reducer. The framework and workflow of FIT-RAG are illustrated in Figure~\ref{fig:overview} and Algorithm~\ref{alg:infer}  respectively.

FIT-RAG comprises five components: (1) a \textbf{Similarity-based Retriever} which selects amount of candidate documents from the corpus; (2) a \textbf{Bi-label Document Scorer} which scores the  candidate documents based on both the factual information and LLM preferences; (3) a \textbf{Bi-faceted Self-Knowledge Recognizer} which determines if external knowledge is necessary for the given question; (4) a \textbf{Sub-document-level Token Reducer} which selects the Top-10 candidate documents and compresses them by extracting the sub-documents;  (5) a \textbf{Prompt Construction} which constructs a prompt based on the question, results of the self-knowledge recognizer and the output of the token reducer.

When a question comes, FIT-RAG first adopts the similarity-based retriever to obtain the 100 most relevant candidate documents based on vector similarity. Here, we build the similarity-based retriever based on Contriever~\cite{contriever}. Next, the bi-label document scorer is performed to score the candidate documents . The scores include the factual information label score (\textit{Has\_Answer}) and the LLM preference label score (\textit{LLM\_Prefer}), which are given based on our proposed bi-label document scorer. Then, the self-knowledge recognizer evaluates whether external knowledge is necessary for each question according to whether the question is related to long-tail or out-of-date knowledge and whether the question’s nearest neighbors have self-knowledge. If necessary, the candidate documents are inputted into the token reducer, where the Top-10 candidate documents are selected according to the scores of the bi-label document scorer and then are compressed to a set of sub-documents to reduce the number of tokens; after that, the selected sub-documents and the  question are jointly integrated  into a prompt. Otherwise, if the external knowledge is unnecessary, only the   question is integrated  into a prompt.  Finally, the  prompt is fed to the LLM to generate an answer. Benefiting from the proposed  bi-label document scorer, bi-faceted self-knowledge recognizer, and sub-document-level token reducer,  FIT-RAG  can provide effective and efficient augmentation for the LLMs. In the following sections, we introduce the details of these novel components.

\subsection{Bi-Label Document Scorer}
\label{sec:scorer}

To evaluate both the factual information and LLM's preference for a retrieved document, we propose to learn a bi-label document scorer using bi-label learning, where the two labels are defined as (1) \textit{Has\_Answer} and (2) \textit{LLM\_Prefer}. The factual information label \textit{Has\_Answer} assesses the likelihood of the document containing the answer to the question, while the LLM preference label \textit{LLM\_Prefer} measures the document's usefulness in helping the LLM generate an accurate response. To learn the scorer, we first formulate the bi-label learning problem for learning a bi-label document scorer.  However, there is serious data imbalance occurring in the training data, which degenerates the performance of the bi-label document scorer. To solve the data imbalance, we propose a data-imbalance-aware bi-label learning algorithm for the bi-label document scorer.

\subsubsection{Bi-Label Learning for Bi-Label Document Scorer}
Consider a bi-label classification problem over an input space $\mathcal{X}$ and a output space $\mathcal{Y}=\{0,1\}^2$. Given a  training data \(D\),  we learn a model \(h(\cdot, \theta):\mathcal{X} \rightarrow \mathcal{Y}\) from a parameterized hypothesis class $\mathcal{H}$, where \(\theta\) represents the model parameters.
For a training sample $ (x, y)\in D$, $y = (y_{1}, y_{2})$ where $y_1$ and $y_2$ represent the \textit{Has\_Answer} label and  \textit{LLM\_Prefer} label respectively. The loss function is denoted by \(l(\cdot, \cdot)\) , and we use binary cross-entropy loss as the loss function, that is $l(h(x,\theta), y) = -\sum_{i=1}^{2} y_{i} \log(h_{i}(x,\theta)) + (1-y_{i})\log(1-h_{i}(x,\theta))$.

We learn the  bi-label scorer by optimizing the following learning objective. 
\begin{equation}
\label{eq: loss_ori}
    \mathop{\min}\limits_{\theta}  L(\theta, D) = \frac{1}{|D|} \sum_{(x,y)\in D} l(h(x,\theta),y)
\end{equation}

\begin{figure}[t]
  \centering
  \includegraphics[width=\linewidth]{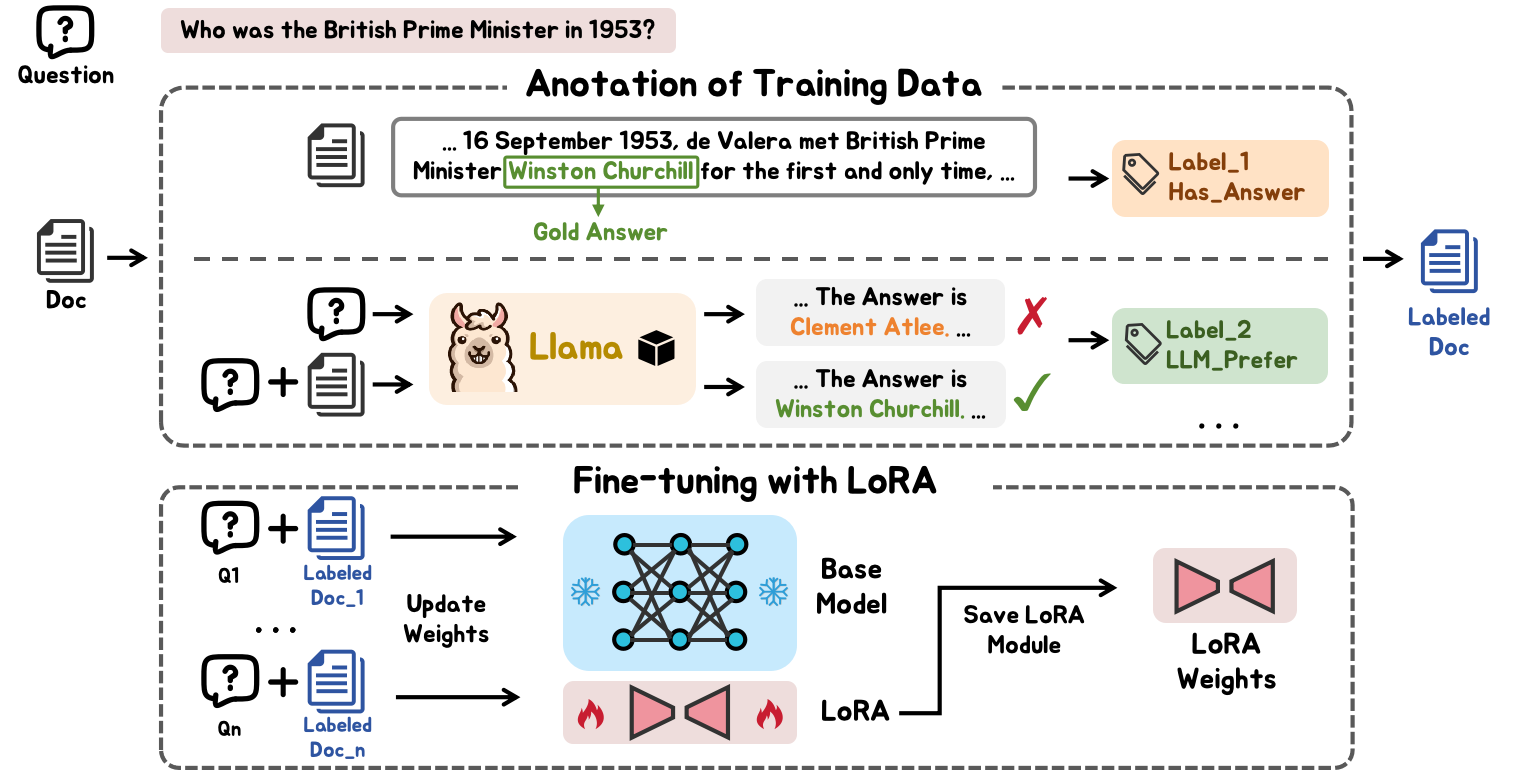}
  \caption{The training process of the Bi-Label Document Scorer}
  \label{fig:bi-label train}
  \Description{}
\end{figure}

To learn the bi-label document scorer, we collect the training set \(D\) and build the classification model $h$ as follows. The process of training data annotation and model fine-tuning is demonstrated in Figure \ref{fig:bi-label train}.

\textbf{The training set} consists of the Top-50 documents for each question that are retrieved from the corpus using Contriever~\cite{contriever}. They are annotated with the following principles. For the \textit{Has\_Answer} label, we annotate it by determining whether the document contains the candidate answers in the gold answer set. If it contains, \textit{Has\_Answer} is labeled as 1, otherwise 0. As to the \textit{LLM\_Prefer} label, we append the document to the question and feed it to the LLM to obtain a predicted answer. If appending the document improves the LLM's performance (i.e., leading to a correct answer), \textit{LLM\_Prefer} is labeled as 1; otherwise, labeled as 0.

\textbf{The classification model} is constructed based on T5~\cite{t5}, where the decoder is replaced by a binary classification head. Furthermore, to accelerate training and save computational resources, we use LoRA  to fine-tune the T5 model.

However, through experimentally analyzing, we find that  the number of data that has $\{1,1\}$ or $\{0,0\}$ labels is typically  more than ten times larger than that for data that has $\{0,1\}$ or $\{1,0\}$ labels. It brings serious data imbalance for this bi-label learning problem and degenerates the performance of the scorer. To address this problem, we propose a data-imbalance-aware bi-label learning algorithm in the next section. 

\subsubsection{Data-imbalance-aware Bi-label Learning} To alleviate the impact of data imbalance, we propose to give different weights for  different data and automatically learn the weights with hypergradient-descent \cite{book/Almeida,conf/iclr/BaydinCMSW18}. The weighted loss is given as follows.
\begin{equation}
     L(\theta, D) = \frac{1}{|D|} \sum_{(x,y)\in D} [f(w)\cdot l(h(x,\theta),y)]
\end{equation}
where
\begin{equation}
     f(w)= w \delta_{y_1, y_2} + ( 1 - w )(1 - \delta_{y_1, y_2}), (y_1, y_2)\in y.
\end{equation}
In the above equation, $w$ is the weight, and  \(\delta_{y_{1}, y_{2}} \) is the Kronecker delta  which is 1 when \(y_{1}\) is equal to \(y_{2}\) and 0 otherwise. To minimize $L(\theta, D)$, we adopt gradient descent. In the $k^{th}$ iteration, $\theta_k$ updates as follows. 
\begin{equation}
    \theta_k = \theta_{k-1} - \eta \nabla L(\theta_{k-1}, D)
\end{equation}
where $\eta$ is the learning rate.

We propose to find $w$ that optimizes the generalization performance of the model by means of hypergradient-descent. Specifically, we randomly split the training data into two subsets: training set $D_t$ and validation set $D_v$, where $D_t$ is used to train the classification model, and $D_v$ is used to estimate generalization loss. The $D_t$ and  $D_v$ is further divided into two subsets respectively according to the values of the two labels: matched train set ($D^{mat}_t$) and matched validation set ($D^{mat}_v$), and mismatched train set ($D^{mis}_t$) and mismatched validation set $D^{mis}_v$. Based on this, the training loss can be defined as
\begin{equation}
    L_t(\theta, D_t) =  w L_{t}(\theta, D_{t}^{mat}) + (1-w)  L_{t}(\theta, D_{t}^{mis}),
\end{equation}
where $L_{t}(\theta, D_{t}^{mat}) = \frac{1}{|D_t|} \sum_{(x,y)\in D_t^{mat}}   l(h(x,\theta),y)$ and $  L_{t}(\theta, D_{t}^{mis}) = \frac{1}{|D_t|} \sum_{(x,y)\in D_t^{mis}}  l(h(x,\theta),y)$,
and the two validation loss can be defined as 
\begin{equation}
    L^{mat}_v(\theta, D^{mat}_v) = \frac{1}{|D^{mat}_v|} \sum_{(x,y)\in D^{mat}_v} l(h(x,\theta),y),
\end{equation}
\begin{equation}
    L^{mis}_v(\theta, D^{mis}_v) = \frac{1}{|D^{mis}_v|} \sum_{(x,y)\in D^{mis}_v}  l(h(x,\theta),y),
\end{equation}
respectively.
Then we formulate the optimization objective as follows:
\begin{equation}
\label{eq:obj}
\begin{split}
    &\min_w \left( L^{mat}_v(\theta_k, D^{mat}_v), \ L^{mis}_v(\theta_k, D^{mis}_v)\right) \\
    &\text{s.t.} \quad \theta_k = \theta_{k-1} - \eta  \nabla_{\theta} L_{t}(\theta_{k-1}, D_{t})
\end{split}
\end{equation}
Based on Equation(\ref{eq:obj}), we first find the hypergradient direction w.r.t w for the label matched data and the label mismatched data:
\begin{equation}
\begin{split}
    d_{mat} = \frac{\partial L^{mat}_v(\theta_{k}, D^{mat}_v)}{\partial w} = \eta \nabla_{\theta} L^{mat}_v(\theta_{k}, D^{mat}_v) \cdot (\nabla_{\theta} L_t(\theta_{k-1}, D_t^{mat}) - \nabla_{\theta} L_t(\theta_{k-1}, D_t^{mis})).
\end{split}
\end{equation}
\begin{equation}
\begin{split}
    d_{mis} = \frac{\partial L^{mis}_v(\theta_{k}, D^{mis}_v)}{\partial w} = \eta  \nabla_{\theta} L^{mis}_v(\theta_{k}, D^{mis}_v) \cdot (\nabla_{\theta} L_t(\theta_{k-1}, D_t^{mat}) - \nabla_{\theta} L_t(\theta_{k-1}, D_t^{mis})).
\end{split}
\end{equation}
By uniformly summing these two direction, we expect to obtain a common descent direction. Define this common direction as $d_{com} = (d_{mat} + d_{mis})/2$. $w$ can be updated as follows.
\begin{equation}
    w_{k} = w_{k-1} - \alpha d_{com}
\end{equation}
where $\alpha$ is the hypergradient step size.

Overall, we propose the data-imbalance-aware bi-label learning algorithm, which is presented in algorithmic form in Algorithm \ref{alg:HD}.

\begin{algorithm}[t]
\caption{Data-imbalance-aware Bi-label Learning Algorithm}
\label{alg:HD}
\SetKwFunction{FMain}{Main}
\SetKwFunction{FGetDirection}{GetDirection}

\textbf{Input:} Training set $\mathcal{D}_t$, validation set $\mathcal{D}_v$, step size $\alpha$ for updating $w$, learning rate $\eta$ for updating the model.\\
\textbf{Initialize:} $w_0$, $\theta_0$ \\

\For{$k=1$ to $K$} {
    $\theta_k \gets \theta_{k-1} - \eta \nabla_{\theta} L_{t}(\theta_{k-1}, D_{t})$\\
    $d_{mat} \gets \eta \nabla_{\theta} L^{mat}_v(\theta_{k}, D^{mat}_v) \cdot (\nabla_{\theta} L_t(\theta_{k-1}, D_t^{mat}) - \nabla_{\theta} L_t(\theta_{k-1}, D_t^{mis}))$\\
    $d_{mis} \gets \eta \nabla_{\theta} L^{mis}_v(\theta_{k}, D^{mis}_v) \cdot (\nabla_{\theta} L_t(\theta_{k-1}, D_t^{mat}) - \nabla_{\theta} L_t(\theta_{k-1}, D_t^{mis}))$\\
    $d_{com} = (d_{mat}+d_{mis}) / 2 $ \\
    $w_{k} = w_{k-1} - \alpha d_{com}$\\
}
\end{algorithm}

\subsection{Bi-faceted Self-Knowledge Recognizer}
\label{sec:recognizer}

\begin{figure}[t]
  \centering
  \includegraphics[width=0.96\linewidth]{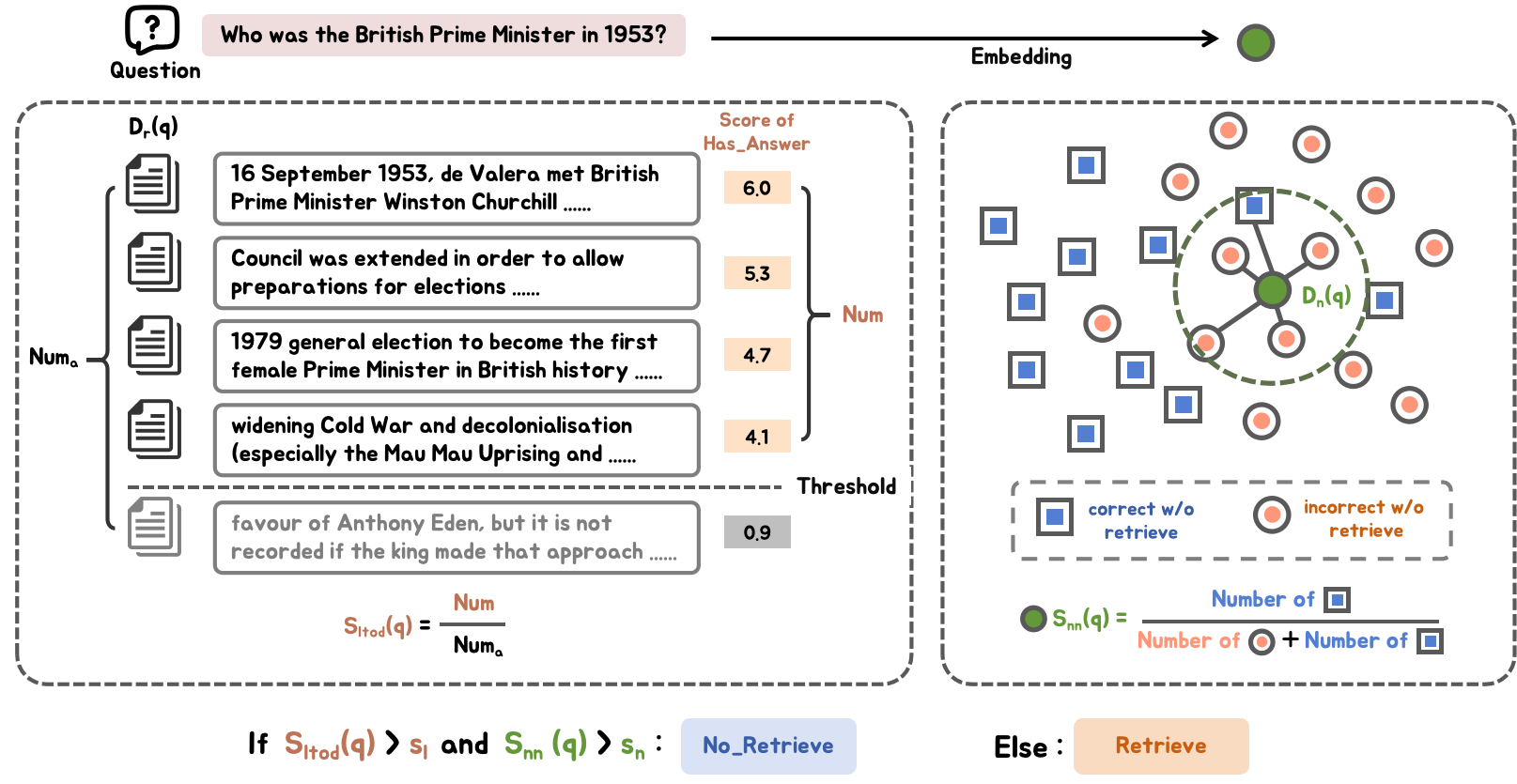}
  \caption{The inference process of Bi-faceted Self-Knowledge Recognizer.}
  \label{fig:recognizer}
\end{figure}

Given a question $q$, we determine whether retrieval is necessary by recognizing whether the LLM has self-knowledge on this question, namely whether the LLM can answer this question without retrieving 
external documents. This paper determines whether the LLM has the self-knowledge based on two facets: (1) whether the question is related to long-tail knowledge or out-of-date knowledge; (2) whether the question's nearest neighbors have self-knowledge. We illustrate the inference process of bi-faceted self-knowledge recognizer in Figure~\ref{fig:recognizer}.

To detect whether the question is related to long-tail knowledge or out-of-date knowledge, we need access to the pretraining corpus and memorization in LLMs; however, they are unavailable for black-box LLMs. To tackle this problem, existing work~\cite{popqa} utilizes Wikipedia's monthly page views as a proxy to simulate the pretraining corpus and achieves proper performance. Following this idea, this paper utilizes Wikipedia's monthly page views as a proxy and determines whether the question is related to long-tail knowledge or out-of-date knowledge based on it.  Based on the output of the retriever $D_r(q)$,  we adopt the \textit{Has\_Answer} label of the Bi-Label Document Scorer and define a score to measure the degree of the question's relevance to the long-tail knowledge or out-of-date knowledge. The score is defined as follows.

\begin{equation}
    S_{ltod}(q) = \frac{1}{|D_r(q)|}\sum_{x \in D_r(q)} \mathbbm{1}_{[h_{ans}(x,\theta) > \delta_{ltod}]}(x), 
\end{equation}

\noindent where $\mathbbm{1}_{[h_{ans}(x,\theta) > \delta_{ltod}]}(x)$ is an indicator function which equals to 1 if $h_{ans}(x,\theta) > \delta_{ltod}$ otherwise 0.  $h_{ans}(x,\theta)$ is the output of Bi-Label Document Scorer on the \textit{Has\_Answer} label, and $\delta_{ltod}$ is a hyper parameter.

Besides, the self-knowledge of the question's nearest neighbors is an important facet for recognizing whether the LLM has self-knowledge on the given question \cite{self-knowledge}. In this paper, we first label a set of questions in the training set as \textit{correct\_w/o\_retrieve} or \textit{incorrect\_w/o\_retrieve} based on whether the LLM can directly answer correctly, and then transform the questions into embedding space by using the T5 encoder and assess their similarity by measuring the Euclidean distance between them. For a given question, this paper finds k nearest neighbors for it. The set of nearest neighbors is denoted as $D_{n}(q)$. Based on $D_{n}(q)$, we design a score to measure self-knowledge of the question's nearest neighbors for the given question. The score is defined as follows.

\begin{equation}
    S_{nn}(q) = \frac{1}{|D_{n}(q)|}\sum_{x \in D_{n}(q)} \mathbbm{1}_{[l_{x}= correct\_w/o\_retrieve]}(x), 
\end{equation}

\noindent where $\mathbbm{1}_{[l_{x}= correct\_w/o\_retrieve]}(x)$ is an indicator function which equals to 1 if the label of x is \textit{correct\_w/o\_retrieve} otherwise 0.  $l_x$ is the label of the corresponding question. 

Combining the above two facets, this paper constructs a bi-faceted self-knowledge recognizer $\mathcal{K}(q)$ as follows.

\begin{equation}
    \mathcal{K}(q) =
    \begin{cases}
        \textit{No\_Retrieve}, & \text{if } S_{ltod}(q) > s_l \text{ and } S_{nn}(q) > s_n, \\
        \textit{Retrieve}, & \text{otherwise},
    \end{cases}
\end{equation}
where \textit{Retrieve} means the given question requires Retrieval and vice versa. $s_l$ and $s_n$ are hyperparameters.

\subsection{Sub-document-level Token Reducer}
\label{sec:compressor}

\begin{figure}[h]
  \centering
  \includegraphics[width=\linewidth]{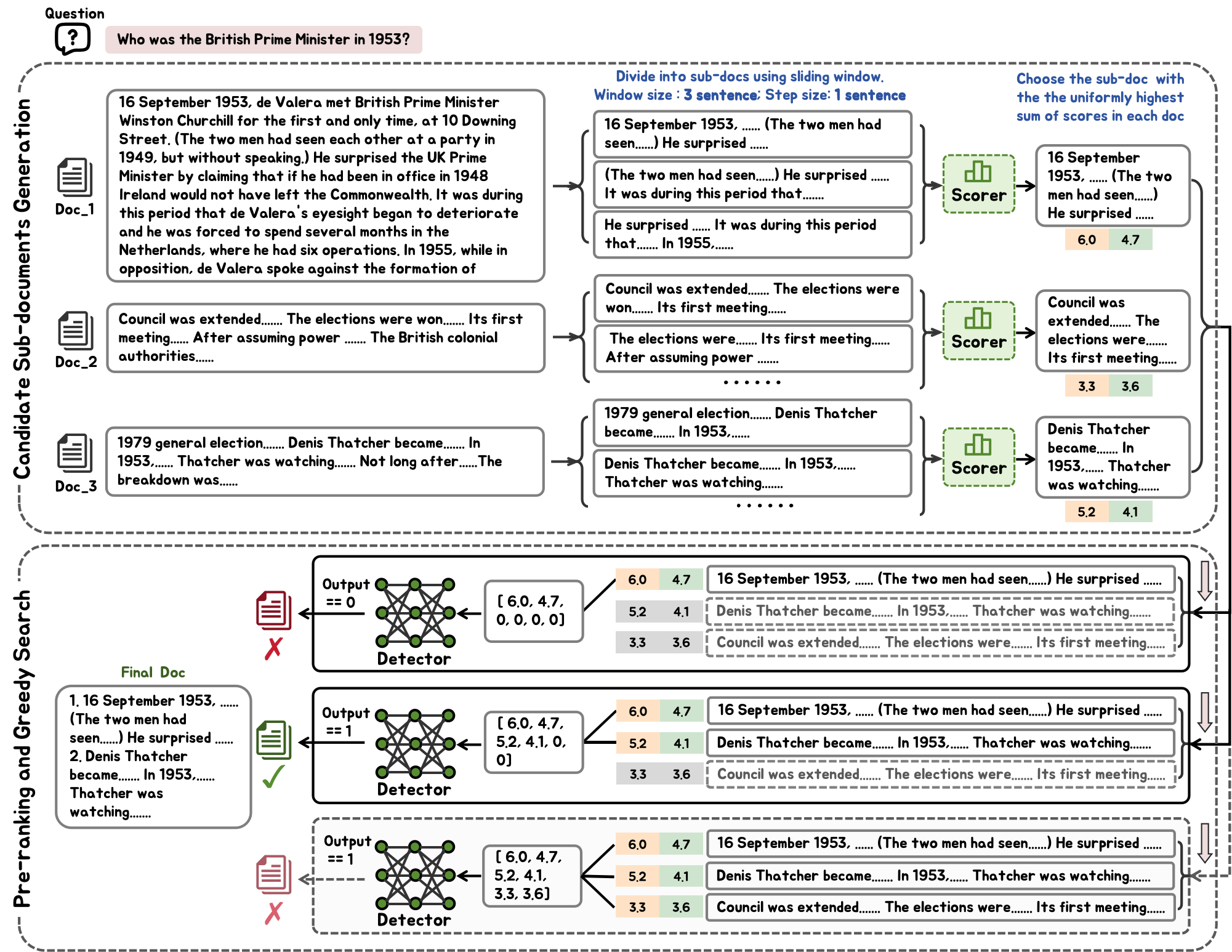}
  \caption{The inference process of Sub-document-level Token Reducer. Here we take three documents for the question as an example.}
  \label{fig:compressor}
\end{figure}

In the \textit{Retrieve} case,  we first rerank the candidate documents using a bi-criteria reranker and select the Top-10 documents. Then, we further  eliminate the redundant tokens by adopting our proposed sub-document-level token reducer. The details of the reranker and token reducer are introduced as follows.

\subsubsection{Bi-criteria Reranker}

The one hundred documents retrieved by the retriever typically contain lots of redundant documents, which not only increase the cost for input tokens but also may confuse the LLM and degenerate the RAG  performance. This paper proposes to eliminate the redundant documents based on two  criteria, namely \textit{Has\_Answer} and \textit{LLM\_Prefer} scores given by the Bi-Label Document Scorer. Specifically, we score the one hundred documents with the uniformly sum of the \textit{Has\_Answer} and \textit{LLM\_Prefer} scores, and then rank the documents according to the score. Based on this rank, we select the Top-10 documents and input them into the sub-document-level token reducer.

\subsubsection{Sub-document-level Token Reducer} The retrieved documents typically contain redundant content that is irrelevant to the question. Eliminating these redundant content can significantly reduce the number of tokens while does not degenerate the RAG performance. This paper proposes a sub-document-level token reducer, which splits the documents into sub-documents and selects a small amount of sub-documents to augment the LLM. It consists of three components: sub-document generator, eligible augmentation detector and sub-document filter.

\noindent \textbf{Sub-document Generator} splits the retrieved documents into sub-documents. Specifically, we apply a sliding window of size 3 (i.e., containing three sentences) with a stride of 1 (i.e., striding one sentence)  on each document and generate a set of three-sentence-constructed sub-documents for each document.

\noindent \textbf{Eligible Augmentation Detector} is a classifier which can determinate whether a combination of sub-documents is eligible to augment the LLM to give correct answers. To learn the classifier, we construct the training data with the following steps: (1) Data Selection. We first select the questions that require retrieval augmentation to be answered correctly from the training data. For each question, we take its Top-10 retrieved documents and split them into sub-documents using the sub-document generator. Then, we randomly combine the sub-documents to form a set of sub-documents combinations and filter out combinations with high overlap. Subsequently, we score each combination using the Bi-Label Document Scorer and generate two scores and only select  the sub-documents combinations whose scores are located on the skyline.  (2) Feature Engineering. For each combination, we concatenate the two scores of all its sub-documents and pad with zeros at the end to maintain the same total input length; (3) Labeling. We concatenate each sub-documents combination with the question and input it to Llama2-13B-Chat. If Llama generates the correct answer, the sub-documents combination is labeled as 1; otherwise, it is labeled as 0.
We build the classifier with a simple four-layer fully connected neural network and train it with the training data. Then, we obtain our eligible augmentation detector and use it to filter out the unnecessary sub-documents.

\begin{algorithm}[t]
    \caption{Sub-document Filter}
    \label{alg:compressor}

    \KwIn{Reranked document set $\mathcal{D}_s$}
    \KwOut{Optimal combination of sub-documents $\mathcal{D}_f$}
    
    \SetKwFunction{FMain}{Main}
    \SetKwFunction{FGenerateSubDocs}{GenerateSubDocs}
    \SetKwFunction{FPreRank}{PreRank}
    \SetKwFunction{FGreedySearch}{GreedySearch}
    \SetKwFunction{FBiLabelScorer}{BiLabelScorer}
    \SetKwFunction{FBinaryDocDetector}{BinaryDocDetector}
    
    \SetKwProg{Fn}{Function}{:}{}
    \Fn{\FGenerateSubDocs{$\mathcal{D}_s$}}{
        $\mathcal{D}_{rep} \gets \emptyset$ \mycomment{\tcp*{Initialize set for representative sub-documents}}
        \ForEach{$d \in \mathcal{D}_s$}{
            $max\_score \gets -\infty$, $rep\_subdoc \gets \text{null}$\;
            \ForEach{$d_s$ generated by sliding window over $d$}{
                $(score_1,score_2) \gets \FBiLabelScorer(d_s)$\;
                \If{$score_1+score_2 > max\_score$}{
                    $max\_score \gets score_1+score_2$, $rep\_subdoc \gets d_s$\;
                }
            }
            $\mathcal{D}_{rep} \gets \mathcal{D}_{rep} \cup \{rep\_subdoc\}$\;
        }
        \KwRet $\mathcal{D}_{rep}$\;
    }

    \Fn{\FPreRank{$\mathcal{D}_{sub}$}}{
        Sort $\mathcal{D}_{sub}$ in descending order based on the sum of their scores\;
        \KwRet Sorted sub-documents\;
    }

    \Fn{\FGreedySearch{$\mathcal{D}_{sorted}$}}{
        $F \gets \emptyset$, $x \gets \emptyset$ \mycomment{\tcp*{Initialize final set and feature vector}}
        \For{$d \in \mathcal{D}_{sorted}$}{
            $x \gets x \cup \{\text{score}_1(d), \text{score}_2(d)\}$ \mycomment{\tcp*{Accumulate reranked score pairs}}
            $s_i \gets \FBinaryDocDetector(x)$ \mycomment{\tcp*{Use MLP model for document selection}}
            \If{$s_i == 1$}{
                $F \gets F \cup \{d\}$ \mycomment{\tcp*{Add to final set if predicted 1}}
                break;
            }
        }
        \KwRet $F$\;
    }
    $\mathcal{D}_{sub} \gets$ \FGenerateSubDocs{$\mathcal{D}_s$}\;
    $\mathcal{D}_{sorted} \gets$ \FPreRank{$\mathcal{D}_{sub}$}\;
    $\mathcal{D}_f \gets$ \FGreedySearch{$\mathcal{D}_{sorted}$}\;
\end{algorithm}

\noindent \textbf{Sub-document Filter} selects sub-documents combination that has few sub-documents but is eligible to augment the LLM to give correct answers. Its workflow is demonstrated in Figure \ref{fig:compressor} and Algorithm~\ref{alg:compressor} respectively, which illustrates a sub-document filtering case involving three sub-documents. From the figure, we can see that the filtering process has three steps: (1) candidate sub-documents generation, where the Sub-document Generator splits each document into multiple sub-documents. These sub-documents are then scored by the Bi-Label Document Scorer, producing two scores for each. The sub-document with the highest sum of scores is selected to represent the original document; (2) eligibility pre-ranking, where the sub-documents obtained in the above step are ranked in descending order according the sum of their two scores; (3) greedy search, where we search the optimal sub-documents combinations in a greedy manner w.r.t the number of sub-documents. The sub-documents combinations are classified with the eligible augmentation detector.  This process begins with one sub-document case. If the current sub-documents combination can not obtain a positive result from the eligible augmentation detector, the process continues by adding one more sub-document. It stops when  the eligible augmentation detector outputs a positive result.

The sub-document-level token reducer can effectively reduce the tokens. In Section \ref{sec:effect of token reducer}, the experimental results demonstrate that it can reduce approximately half of the tokens.

\subsection{Prompt Construction}
\label{sec:prompt}

\begin{figure}[t]
  \centering
  \includegraphics[width=\linewidth]{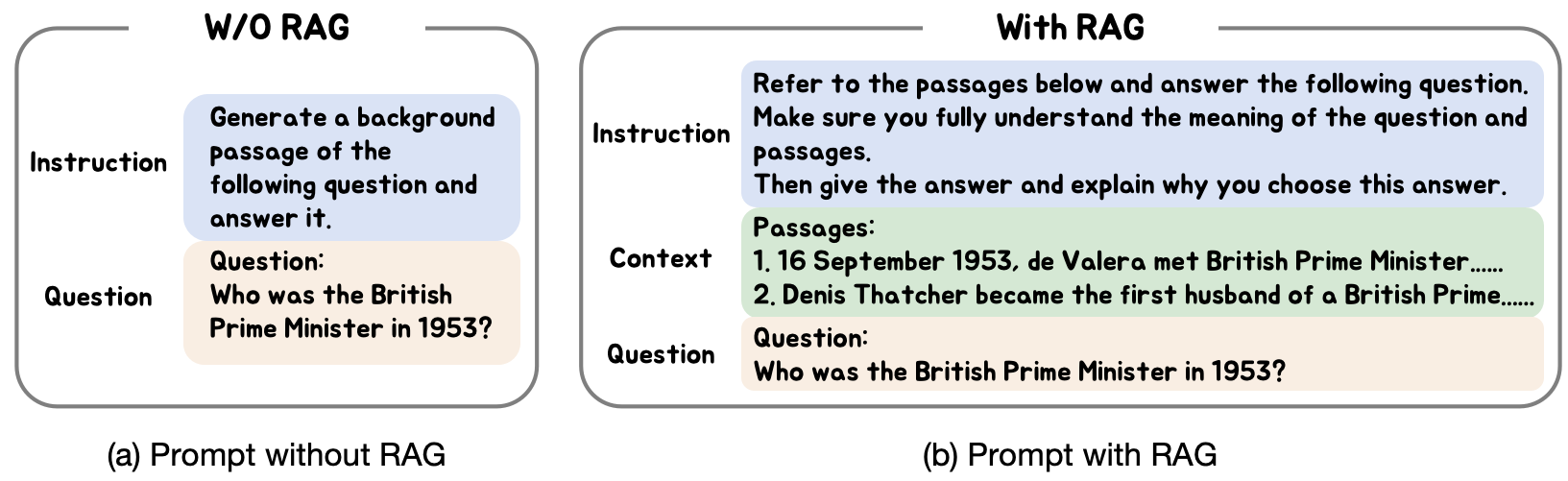}
  \caption{The prompt templates for scenarios with and without RAG}
  \label{fig:prompt}
\end{figure}

This paper adds the retrieved documents into prompts to augment LLMs. The design of the prompt template significantly matters the performance of RAG. This paper designs the  prompt template for the case \textit{Retrieve} as Figure \ref{fig:prompt} (b) shows. In this template, we propose a sophisticated  instruction, which consists of three parts. In the first part, we ask the LLM to \textit{refer to the passage below and answer the following question}, which leads the LLM to answer with the retrieved passages. In the second part, we emphasize that the LLM need to \textit{read and understand the question carefully and make sure it fully understand what the question means}, which excites the LLM's deeper thinking for the question. In the last part, we ask the LLM to \textit{answer the question and explain why you choose this answer}, where asking the LLM to explain why it choose this answer enables LLM to perform better.  Following this instruction, we directly put the retrieved documents as context and then give the question. 

As to the \textit{No\_Retrieve} case,  we follow the idea of provoking the LLM's internal knowledge proposed in GenRead~\cite{genread}. The prompt template is illustrated in Figure \ref{fig:prompt} (a), where we instruct the LLM to first generate a background passage about the question based on its internal knowledge, and then ask the LLM to answer the question by reasoning over its self-generated context. 

%% file: text/5.Experiments.tex
\section{EXPERIMENTS}

\subsection{Experimental Settings}

\subsubsection{Datasets}
Following prior works~\cite{atlas,aar,replug}, we choose TriviaQA~\cite{tqa}, Natural Questions~\cite{nq}, and PopQA~\cite{popqa} as the datasets for our experiments. 
\begin{enumerate}
    \item \textbf{TriviaQA (TQA)} ~\footnote{https://nlp.cs.washington.edu/triviaqa}is a dataset specifically designed for enhancing reading comprehension tasks. It comprises over 650,000 question-answer-evidence combinations. The dataset includes 95K question-answer pairs authored by trivia enthusiasts and independently gathered evidence documents, six per question on average, providing high quality distant supervision for answering the questions.
    
    \item \textbf{Natural Questions (NQ)} ~\footnote{https://ai.google.com/research/NaturalQuestions} is a large dataset for training and evaluating open-domain question-answering systems. It consists of approximately 300,000 question-answer pairs. The questions are genuine, anonymized queries from Google search users, and the answers are drawn from Wikipedia. The dataset requires systems to read and comprehend entire Wikipedia articles for effective answering.
    
    \item \textbf{PopQA} ~\footnote{https://huggingface.co/datasets/akariasai/PopQA} is an open-domain question-answering dataset aimed at assessing the factual knowledge memorization of large language models. Comprising approximately 14,000 questions, it focuses on entities sampled from Wikidata. The dataset emphasizes long-tail knowledge, covering a wide range of factual information often underrepresented in popular QA datasets, making it significant for evaluating language models on less popular factual knowledge.
\end{enumerate}

\subsubsection{Baselines}
We conduct experiments on the TQA, PopQA, and NQ datasets and compare our method with the following baselines:
\begin{enumerate}
    \item \textbf{Llama2-13B-Chat}~\cite{llama2}, the chat-optimized version of Llama2 with around 13 billion parameters. Tailored for conversational tasks, Llama2-13B-Chat is adept at understanding and engaging in dialogues.
    \item \textbf{ChatGPT}~\footnote{https://openai.com/blog/chatgpt}, a variant of the GPT (Generative Pre-trained Transformer) models, which is specifically enhanced for conversational engagement. With its advanced capabilities in processing and generating human-like text, ChatGPT excels in interactive and coherent dialogue creation. 
    \item \textbf{GenRead+Llama2-13B-Chat}, a method that combines the Llama2-13B-Chat model with the GenRead~\cite{genread} method. The GenRead method adopts a generate-then-read approach, which first utilizes the LLM to create contextual documents based on the given question, and then employs the LLM to generate the final answer based on these documents.
    \item \textbf{REFEED+Llama2-13B-Chat}, a method that combines the Llama2-13B-Chat model with the REFEED~\cite{refeed} pipeline. The pipeline first prompts the LLM to generate an initial answer and use a retrieval model to retrieve relevant information based on the question and initial answer. Then it use the retrieved information to help the LLM refine its initial answer.
    \item \textbf{AAR+Llama2-13B-Chat}, a method that augment the Llama2-13B-Chat model with the Augmentation-Adapted Retriever (AAR)~\cite{aar}. AAR use a source language model to provide preference signals and fine-tune the retriever to align with its preferences by utilizing the encoder-decoder attention mechanisms.
\end{enumerate}

\subsubsection{Evaluation Metrics}
Following the evaluation metric used in AAR and GenRead~\cite{aar, genread}, we evaluate the model performance based on whether gold answers are included in the model generations. Furthermore, to evaluate the retrieval performance, we employ Recall@K (R@K) as the evaluation metric, which is the percentage of top-K retrieved documents that contain the answer, following previous work DPR, Contriever, and ANCE~\cite{dpr, ance, contriever}.

\subsubsection{Implementation}
We employ the Wikipedia 2018 dataset~\cite{wiki} as the knowledge corpus, utilize the Llama2-13B-Chat model as our black-box large language model, and  build the similarity-based retriever based on Contriever~\cite{contriever}. The T5-Base~\cite{t5} model is used as the base model to construct the bi-label document scorer; furthermore, the model is fine-tuned with LoRA. Specifically, the rank $r$ of the LoRA matrix is $16$, $lora\_alpha$ is $32$, $lora\_dropout$ is $0.05$. The training set of TriviaQA and NQ is used for training the bi-label document scorer and the Llama2-13B-Chat~\cite{llama} is adapted to label the training data. During training, the batch size is $16$, the learning rate is $3e-4$, and the number of epochs is $20$. In the bi-faceted self-knowledge recognizer, the hyperparameter $\delta_{ltod}$ is $4.5$, $s_l$ is $0.04$, and $s_n$ is $0.67$ for TriviaQA, $0.55$ for NQ and PopQA. In the sub-document-level token reducer, we use the four-layer fully connected neural network as the detector and use the training set of TriviaQA to label the training data. And finally, we utilize the Llama2-13B-Chat model as our black-box large language model. We run all experiments on a single A100 GPU (40G).

\subsection{Overall Performance}

\setlength{\tabcolsep}{1.2mm}{
\begin{table*}[t]
  \caption{Comparison between baselines and our method in terms of answering accuracy on TriviaQA dataset, NQ dataset and PopQA dataset. Input Tokens indicate the average number of input tokens per question. For Llama2-13B-Chat and ChatGPT, we directly input the question and instruct them to give the answer.}
  \label{tab: overall results}
  \small
  \centering
      \begin{tabular}{c|c|cc|cc|cc}
        \toprule
        \multirow{2}{*}{Model} & \multirow{2}{*}{\# Parameters} & \multicolumn{2}{c|}{TriviaQA} & \multicolumn{2}{c|}{NQ} & \multicolumn{2}{c}{PopQA}\\ 
        & & Acc (\%) & Input Tokens & Acc (\%) & Input Tokens & Acc (\%) & Input Tokens \\
        \midrule
        Llama2-13B-Chat & 13B & 60.9 & 47 & 34.1 & 35 & 26.9 & 37  \\
        ChatGPT & $\gg$13B & 69.5 & 47 & 34.5 & 35 & 35.4 & 37  \\
        GenRead + Llama2-13B-Chat & 13B & 68.0 & 952 & 44.3 & 1272 & 34.7 & 1179 \\
        REFEED + Llama2-13B-Chat & 13B & 71.7 & 1793 & 50.5 & 1766 & 42.9 & 1634 \\
        AAR + Llama2-13B-Chat & 13B & 69.8 & 1689 & 47.9 & 1683 & 46.8 & 1540 \\
        \textbf{FIT-RAG + Llama2-13B-Chat (ours)} & 13B & \textbf{75.2} & \textbf{816} & \textbf{54.0} & \textbf{1059} & \textbf{54.4} & \textbf{883} \\
        \bottomrule
      \end{tabular}
\end{table*}}

In Table~\ref{tab: overall results}, we report the performance of our method and baseline methods on TriviaQA dataset, NQ dataset, and PopQA dataset respectively. From this table, we can see that our method clearly outperforms all the baseline methods on the three datasets and consumes the lowest tokens compared with other black-box RAG systems like REFEED and AAR. Specifically, our method achieves 75.2\% in terms of answering accuracy on TriviaQA dataset, 54.0\% on NQ dataset and 54.4\% on PopQA dataset.

Compared with the baseline Llama2-13B-Chat, our method improves the accuracy by 14.3\% on TriviaQA dataset, 19.9\% on NQ dataset and 27.5\% on PopQA dataset, respectively, which demonstrates the significant benefits of external knowledge augmentation. Notably, compared with ChatGPT, which is widely considered to have many more parameters than  Llama2-13B-Chat, our RAG-augmented Llama model is able to surpass the performance of ChatGPT by 5.7\% on the TriviaQA dataset, 19.5\% on NQ dataset and 19\% on PopQA dataset. This highlights that effectively incorporating external knowledge can compensate the model size, allowing even mid-sized LLMs to rival the capabilities of much larger models. Furthermore, we can see that the gap between different datasets varies. On TriviaQA, the performance gap between ChatGPT and Llama2-13B-Chat enhanced with RAG is relatively small. However, on more challenging QA datasets NQ and PopQA, the gaps are substantial. This is because these datasets involve more updated and long-tail knowledge,  making it challenging to rely solely on LLM's inherent knowledge. By effectively augmenting with external information, our method provides greater benefits on such knowledge-intensive datasets, enabling substantial capability enhancement beyond what can be achieved through scaling model parameters alone.

Compared with the state-of-the-art black-box RAG methods,  AAR and REFEED, the improvement in our approach is also significant. Moreover, from Table~\ref{tab: overall results} we can see that our method consumes substantially fewer tokens than the other black-box RAG methods. On average, our approach reduces the number of input tokens per question by approximately half across the three datasets compared with REFEED and AAR. It demonstrates that our approach not only enhances RAG performance but also improves computational efficiency by preventing the LLM from being overloaded with excessive irrelevant tokens.

\subsection{Effect of Bi-label Document Scorer}

In this section, we experimentally investigate the impact of the bi-label document scorer on RAG performance. We first analyze the effects of the two labels (i.e.,  the factual information label (\textit{Has\_Answer}) and the LLM preference label  (\textit{LLM\_Prefer})). Then, we conduct an ablation  study on the data-imbalance-aware bi-label learning algorithm.

\subsubsection{Effect of the Two Labels} 

To explore the effect of each label, we evaluate both the recall of the retrieved documents and the answering accuracy obtained with the retrieved documents. For the recall performance, we use the R@k metric, which calculates the proportion of documents containing the gold answers within the Top-k retrieved documents. Specifically, we first use the  Contriever to retrieve Top-100 documents from the Wikipedia corpus, then score them by \textit{Has\_Answer} score, \textit{LLM\_Prefer} score and uniformly sum of the two scores, respectively. Based on the scores, we  rerank the Top-100 documents and select the Top-k from them.  For the answering accuracy, we record the accuracy of answers generated by Llama2-13B-Chat when using the Top-10 reranked documents as the external knowledge. For convenience, the token reducer is not involved, and the Top-10 documents are directly integrated into the prompt template shown as  Fig~\ref{fig:prompt} (b). Then, we feed the prompt into Llama2-13B-Chat to generate answers.

\begin{figure}[t]
  \centering
  \includegraphics[width=0.96\linewidth]{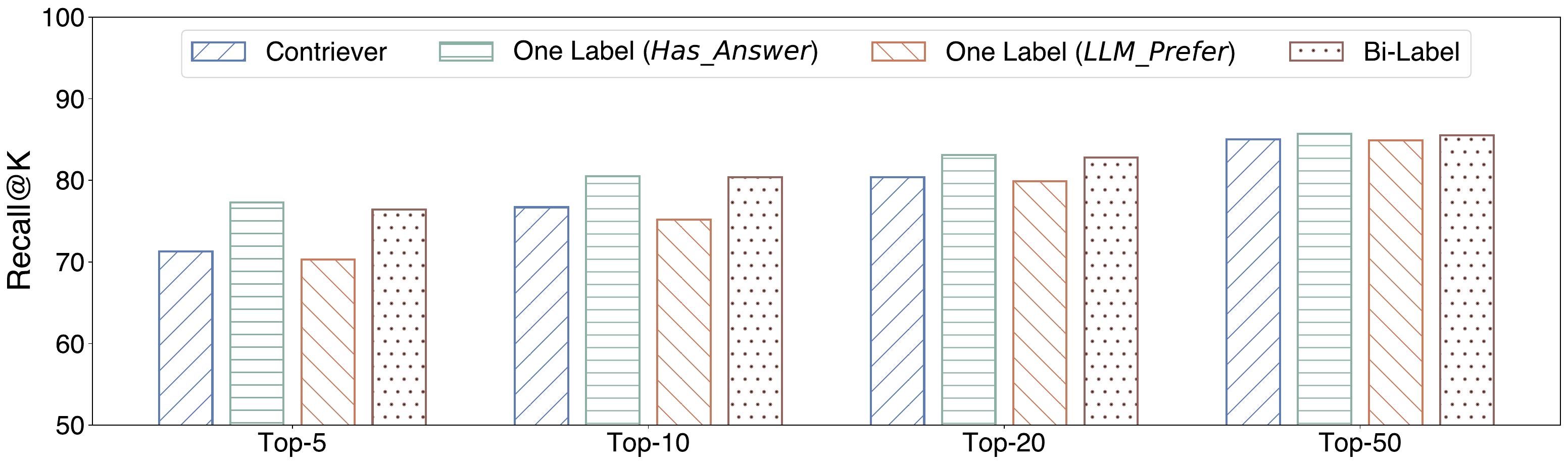}
  \caption{The recall@k of the reranked top-100 documents on TriviaQA dataset}
  \label{fig:tqa recall}
\end{figure}

\begin{figure}[t]
  \centering
  \includegraphics[width=0.96\linewidth]{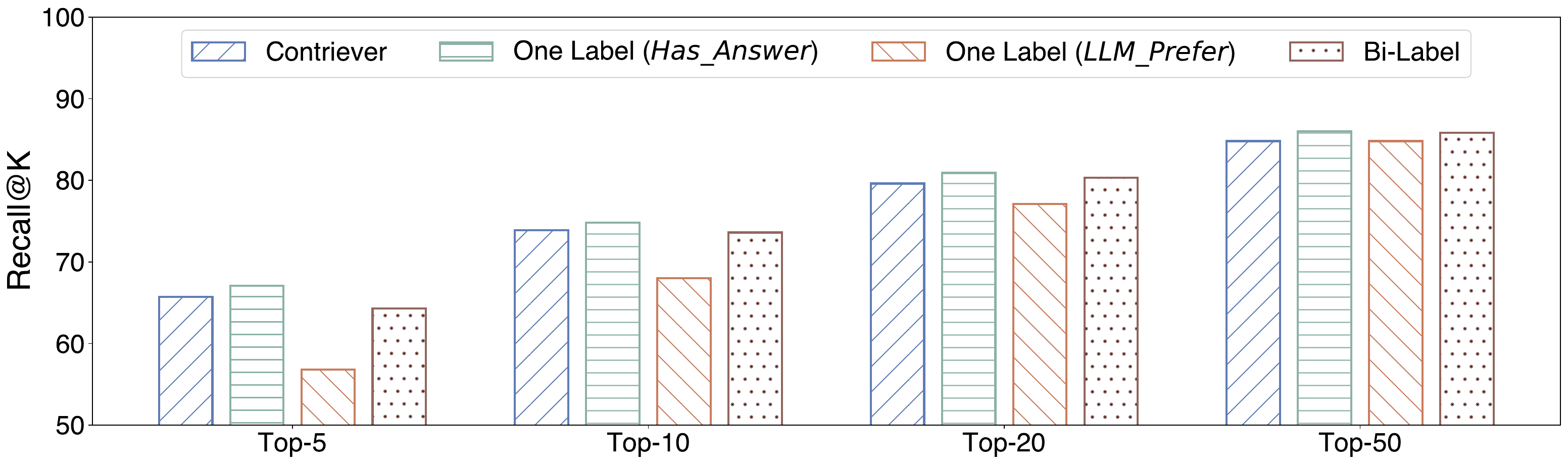}
  \caption{The recall@k of the reranked top-100 documents on NQ dataset}
  \label{fig:nq recall}
\end{figure}

\begin{figure}[t]
  \centering
  \includegraphics[width=0.96\linewidth]{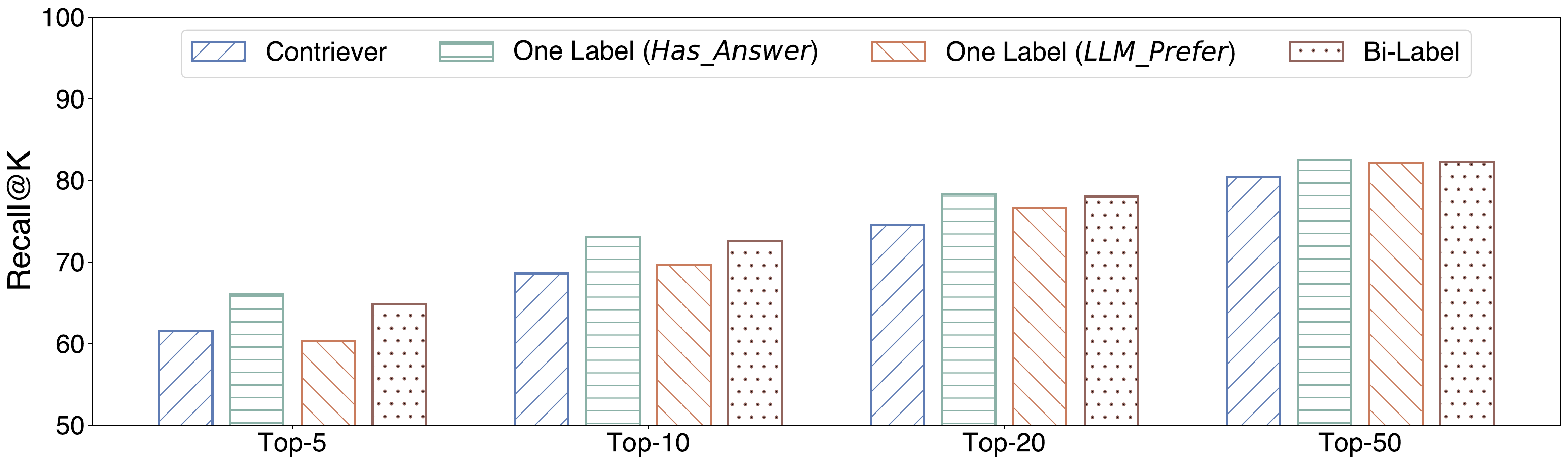}
  \caption{The recall@k of the reranked top-100 documents on PopQA dataset}
  \label{fig:pqa recall}
\end{figure}

\begin{figure}[t]
  \centering
  \includegraphics[width=0.96\linewidth]{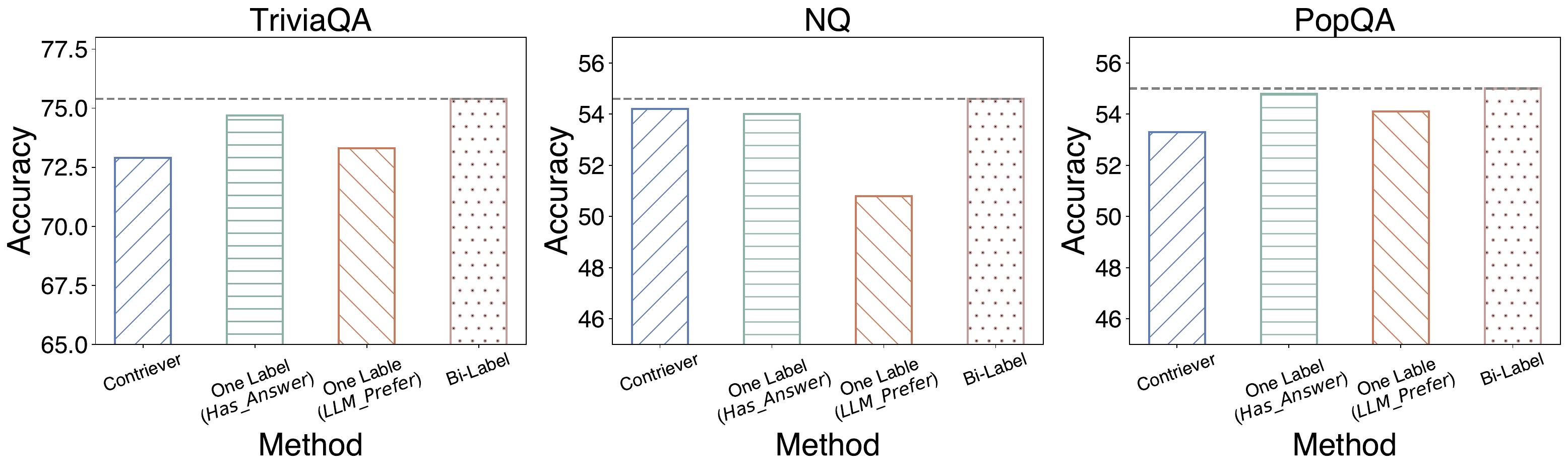}
  \caption{Comparison between the answering accuracy achieved by  contriever, \textit{Has\_Answer} score-based rerank, \textit{LLM\_Prefer}  score-based rerank, and bi-label rerank, where contriever represents the method that does not involve  reranking. }
  \label{fig:acc}
\end{figure}

The results of R@k on three datasets are recorded in Figure~\ref{fig:tqa recall} to Figure~\ref{fig:pqa recall}, respectively, and the answering accuracy is recorded in Figure~\ref{fig:acc}. We can see that \textit{Has\_Answer} score-based augmentation has the highest R@k on all the three datasets, and it has higher answering accuracy than the \textit{LLM\_Prefer} score-based augmentation. This result validates the critical importance of factual information that the \textit{Has\_Answer} indicated. As for the label \textit{LLM\_Prefer}, only using the \textit{LLM\_Prefer} score has lower R@k compared to that for only using label \textit{Has\_Answer}. For uniformly sum of the two scores, which considers  \textit{Has\_Answer} and \textit{LLM\_Prefer} together, does not improve the R@k, but it significantly improves the answering accuracy.
This is because  \textit{Has\_Answer} ensures documents contain factual information, and  \textit{LLM\_Prefer} selects documents which are useful to the LLM. Combining the two labels can provide documents that contain factual information and are LLMs preferred, which  improve the answering accuracy. We can see from Figure~\ref{fig:acc} that our proposed bi-label scorer improves the accuracy by 2.5\% in TriviaQA, 0.4\% in NQ and 1.7\% in PopQA compared with the similarity-based retriever, showing the effectiveness of our method. 

\subsubsection{Ablation Study on the Data-imbalance-aware Bi-label Learning Algorithm.}
\begin{figure}[b]
  \centering
  \includegraphics[width=0.5\linewidth]{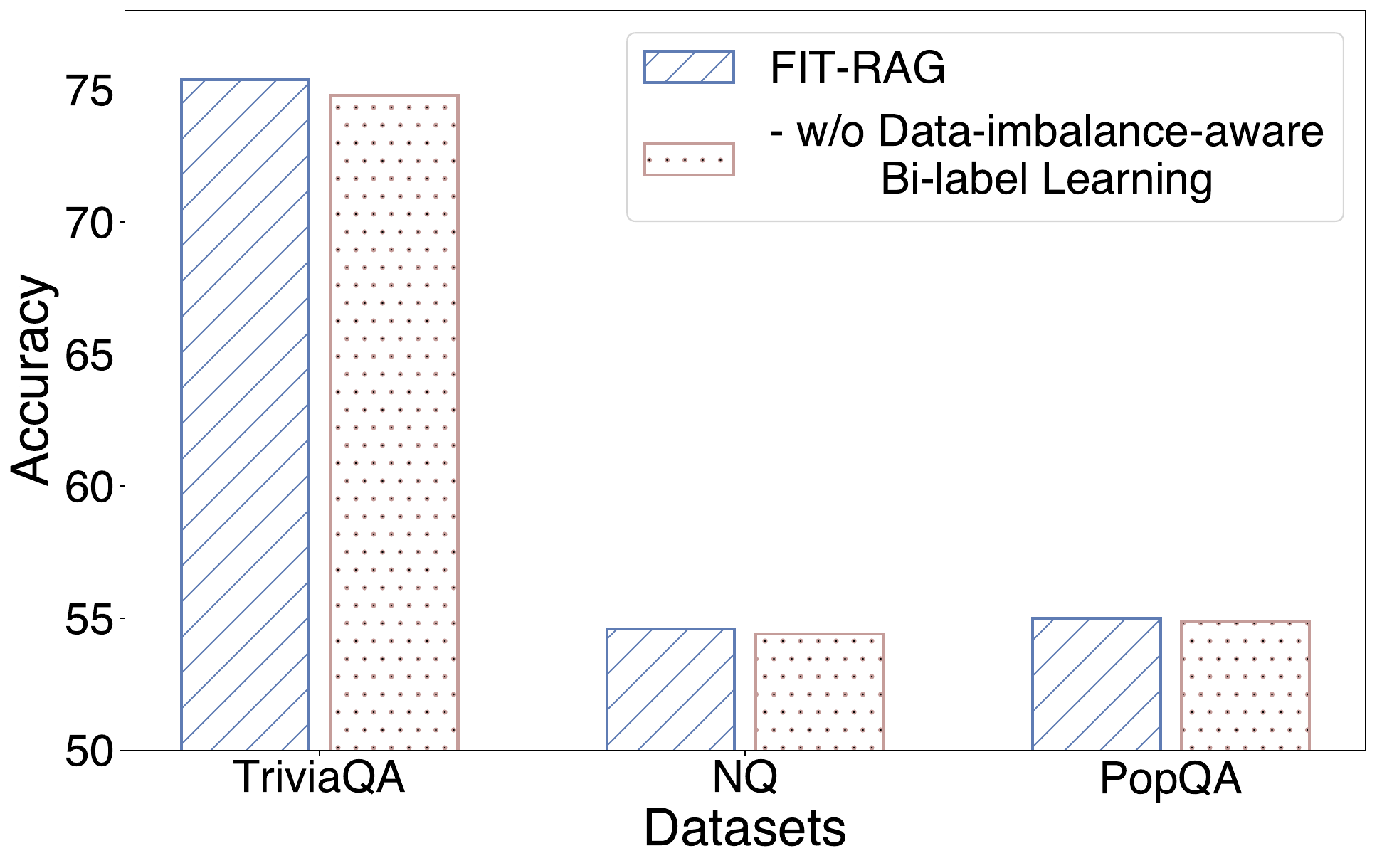}
  \caption{Comparison between the answering accuracy of with/without data-imbalance-aware bi-label learning algorithm.}
  \label{fig:imbalance}
\end{figure}

To investigate the effect of data-imbalance-aware bi-label learning algorithm, we conduct ablation study on it.
For convenience, the token reducer is not involved, and the Top-10 documents are directly integrated into the prompt template shown as  Fig~\ref{fig:prompt} (b). Then we use Llama2-13B-Chat to generate the answers according to the input prompt. The results are recorded in Figure~\ref{fig:imbalance}.  From this figure, we can see that the answering accuracy drops by 0.6\% on TriviaQA, 0.2\% on NQ and 0.1\% on PopQA respectively, when the data-imbalance-aware bi-label learning algorithm is excluded. The drops on the  NQ and  PopQA are not significant because of the  labels in these two datasets is highly noisy and the imbalance ratio is undetermined.  This result demonstrates that our approach can effectively address data imbalance and improve the performance of black-box RAG. 

\subsection{Effect of Bi-faceted Self-Knowledge Recognizer}

\begin{figure}[t]
  \centering
  \includegraphics[width=0.96\linewidth]{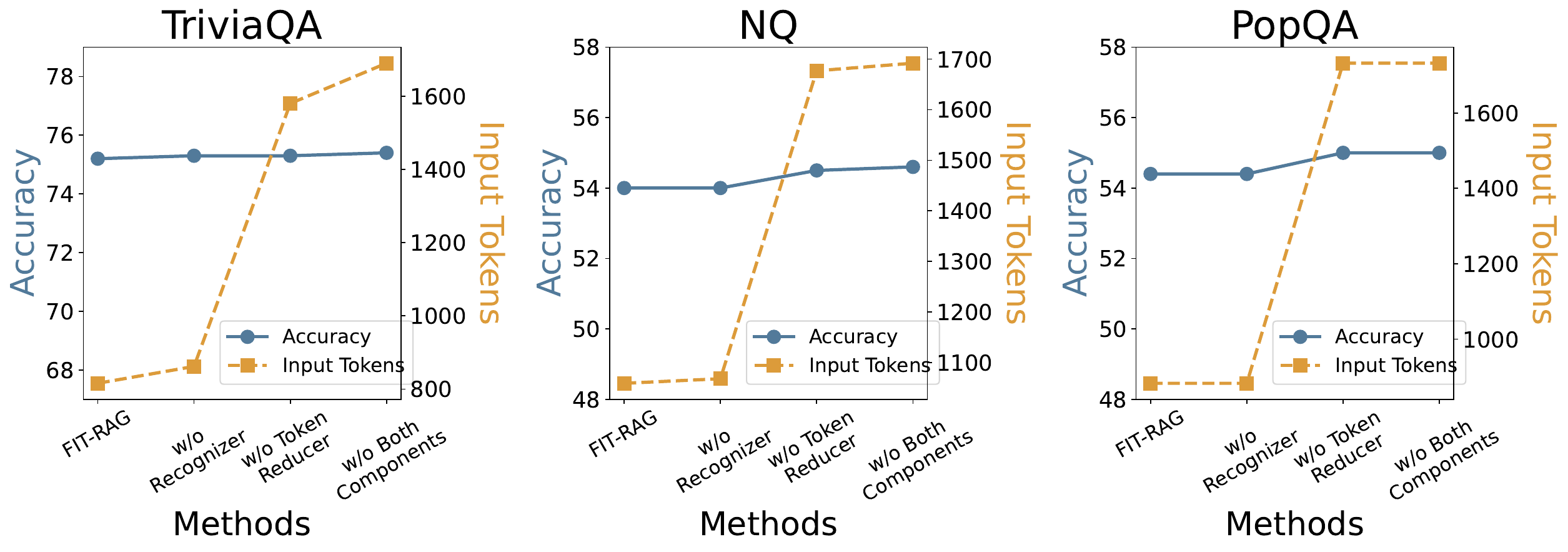}
  \caption{The impact of bi-faceted self-knowledge recognizer and sub-document-level token reducer on the accuracy of answers and the average input tokens per question.}
  \label{fig:token}
\end{figure}

To investigate the effect of the bi-faceted self-knowledge recognizer, we conduct an ablation study on it. The results are recorded in Figure~\ref{fig:token}. From this figure, we can see that our proposed  bi-faceted self-knowledge recognizer can significantly reduce the number of tokens while does not decrease the answering accuracy for the TriviaQA dataset. By contrast, the token reduction effects on the NQ and PopQA datasets are not substantial. This is because that  Llama2-13B-Chat has less self-knowledge for these two datasets and requires retrieval for most of the questions in these two datasets. Our bi-faceted self-knowledge recognizer reduce input tokens by reducing unnecessary retrieval. For the datasets that requires retrieval for most of the questions, the effect of token reduction of our proposed recognizer is limited.

\subsection{Effect of Sub-Document-Level Token Reducer}
\label{sec:effect of token reducer}

In this section, we investigate the effect of sub-document-level token reducer. We first conduct an ablation study on it. The results are illustrated in Figure~\ref{fig:token}. From this figure, we can see that our proposed token reducer can significantly reduce the input tokens while not decreasing the answering accuracy. Specifically, compared with the original input tokens, our method can reduce 49\% of input tokens on the TriviaQA dataset, 37\% on the NQ dataset and 49\% on the PopQA dataset. The results demonstrate the effectiveness of our proposed token reducer.

\begin{figure}[t]
  \centering
  \includegraphics[width=0.96\linewidth]{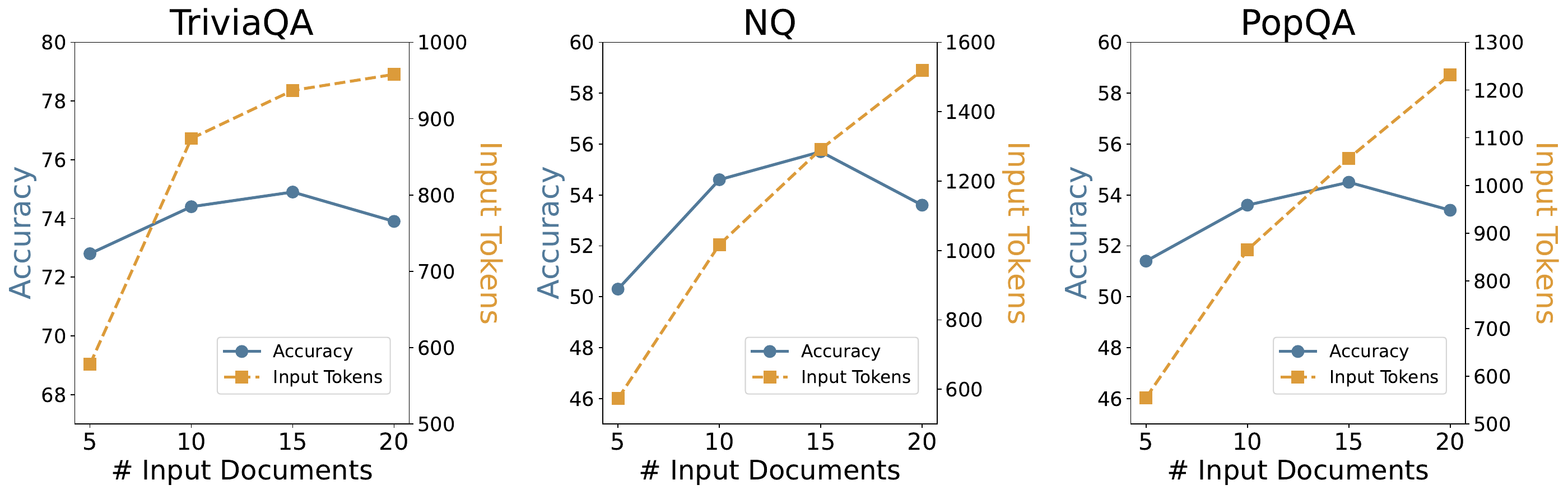}
  \caption{The accuracy of answers and the average input tokens per question when choosing top-k documents that input to the token reducer. For each dataset, we randomly choose 3000 samples for experiment.}
  \label{fig:topk}
\end{figure}

Next, we investigate the impact of the  number of documents that are inputted to the token reducer.  Specifically, we set the number of documents to be 5, 10, 15, and 20 respectively and observe the changes of the RAG performance. For each document, we choose the sub-document with the highest score that obtained by uniformly sum of two scores generated by the bi-label document scorer, and add it to the candidate sub-documents list. Then we use the sub-document filter to choose the final sub-document combinations as the external knowledge for the LLM. We report the changes of RAG performance in Figure~\ref{fig:topk}. From this figure, we can see that as the number of documents increases, the number of input tokens also rises. When 5 documents are inputted, the model has the lowest answering accuracy. This demonstrates that too few input documents restrict the amount of knowledge available for augmentation and degenerates the answering accuracy. With the number of input documents increasing to 10, we observe a corresponding improvement w.r.t answering accuracy. However, as the number of input documents reaches 20, there is a decline in the  answering accuracy, which may be cause by the involving of redundant information. Overall, we can see that setting the number as 10 can achieve a proper trade-off between answering accuracy and the effect of token reduction, and we use it in our sub-document-level token reducer.

\subsection{Effect of Prompt Construction}
\label{sec:effect_prompt}

In this section, we experimentally compare the performance of different prompt templates.
As shown in Table~\ref{tab: prompt}, we conduct experiments with three types of prompts: (1) Simple Prompt. This basic prompt just simply concatenates a simple instruction and the augmentation documents provided by FIT-RAG; (2) CoT Prompt. This prompt is based on the Chain-of-Thought (CoT) prompting method~\cite{cot}, which guides the model to reason step by step. Specifically, we add the CoT prompt \textit{Let's think step by step} at the end of the  simple prompt; (3) Comprehensive Prompt. This is  our proposed sophisticated prompt used in scenarios where retrieval is needed, as introduced in Section~\ref{sec:prompt}.
The answering accuracy w.r.t different prompts are recorded in Table~\ref{tab: prompt}. From this table, we can see that our proposed prompt outperforms the simple prompt and CoT prompt by 2.7\% and 1.5\%, respectively. It demonstrates that our proposed prompt can help to achieve proper RAG performance.

\begin{table}[tbp]
  \caption{Comparison between different prompt templates.}
  \label{tab: prompt}
  \begin{tabular}{c|c|c}
    \toprule
    Prompt Name & Prompt Template & Acc \\
    \midrule
    \multirow{6}{*}{Simple Prompt} &  \parbox[t]{9cm}{\textbf{Refer to the passage below and answer the following question.}\\
    Passages: \\
    1. 16 September 1953, de Valera met British Prime Minister...... \\
    2. Denis Thatcher became the first husband of a British Prime...... \\
    Question: Who was the British Prime Minister in 1953? \\
    \textbf{The answer is}} & \multirow{6}{*}{72.7}\\
    \addlinespace
    \hline
    \addlinespace
    \multirow{6}{*}{CoT Prompt} &  \parbox[t]{9cm}{\textbf{Refer to the passage below and answer the following question.}\\
    Passages: \\
    1. 16 September 1953, de Valera met British Prime Minister...... \\
    2. Denis Thatcher became the first husband of a British Prime...... \\
    Question: Who was the British Prime Minister in 1953? \\
    \textbf{Let's think step by step.}} & \multirow{6}{*}{73.9}\\
    \addlinespace
    \hline
    \addlinespace
    \multirow{7}{*}{Comprehensive Prompt (ours)} &  \parbox[t]{9cm}{\textbf{Refer to the passage below and answer the following question.\\
    Make sure you fully understand the meaning of the question\\ and passages.\\
    Then give the answer and explain why you choose this answer.}\\
    Passages: \\
    1. 16 September 1953, de Valera met British Prime Minister...... \\
    2. Denis Thatcher became the first husband of a British Prime...... \\
    Question: Who was the British Prime Minister in 1953?} & \multirow{7}{*}{75.4}\\
    \bottomrule
  \end{tabular}
\end{table}

%% file: text/6.Conclusions.tex
\section{CONCLUSIONS}
In this paper, we propose a novel black-box RAG framework for black-box LLMs, FIT-RAG, which achieves both superior effectiveness and token efficiency. FIT-RAG improves the effectiveness of black-box RAG by utilizing both factual information and LLM preference in retrieval; besides, it boosts the token efficiency of black-box RAG by fully using self-knowledge and conducting  sub-document-level token reduction. With the superior effectiveness and token efficiency, FIT-RAG has the potential to be widely applied in vertical domains. However, this paper only considers the input-augmented RAG mode that inputs the retrieved documents in the prompt. In the future, we will extend FIT-RAG to the output-augmented RAG mode where the retrieved documents are utilized to  edit the output of LLMs.